\documentclass[nohyperref]{article}

\usepackage[dvipsnames]{xcolor}

\usepackage{microtype}
\usepackage{graphicx}
\usepackage{booktabs} %
\usepackage{caption}
\usepackage{subcaption}
\usepackage{dsfont}
\usepackage{longtable}
\usepackage{supertabular}
\usepackage{csquotes}
\usepackage{enumitem}

\usepackage[colorlinks=true,citecolor=MidnightBlue, linkcolor=black, urlcolor=black]{hyperref}

\usepackage[accepted]{icml2023}

\usepackage{amsmath}
\usepackage{amssymb}
\usepackage{mathtools}
\usepackage{amsthm}

\usepackage[capitalize,noabbrev]{cleveref}

\theoremstyle{plain}

\theoremstyle{definition}

\theoremstyle{remark}

\usepackage[textsize=tiny]{todonotes}

\usepackage[most]{tcolorbox}
\usepackage{lipsum}
\usepackage{namedtensor}
\usepackage{soul}
\usepackage{placeins}

\crefformat{equation}{(#2#1#3)}

\usepackage{tikz} %
\usetikzlibrary{arrows, chains, positioning, decorations.pathreplacing, fit, calc, decorations.pathmorphing, patterns, shapes.geometric}
\usepackage{scalerel} %

\definecolor{mplblue}{rgb}{0.12156862745098039, 0.4666666666666667, 0.7058823529411765}
\definecolor{mplorange}{rgb}{1.0, 0.4980392156862745, 0.054901960784313725}
\definecolor{mplgreen}{rgb}{0.17254901960784313, 0.6274509803921569, 0.17254901960784313}
\definecolor{mplred}{rgb}{0.8392156862745098, 0.15294117647058825, 0.1568627450980392}
\definecolor{mplpurple}{rgb}{0.5803921568627451, 0.403921568627451, 0.7411764705882353}
\definecolor{mplbrown}{rgb}{0.5490196078431373, 0.33725490196078434, 0.29411764705882354}
\definecolor{mplpink}{rgb}{0.8901960784313725, 0.4666666666666667, 0.7607843137254902}
\definecolor{mplgray}{rgb}{0.4980392156862745, 0.4980392156862745, 0.4980392156862745}
\definecolor{mplyellow}{rgb}{0.7372549019607844, 0.7411764705882353, 0.13333333333333333}
\definecolor{mplcyan}{rgb}{0.09019607843137255, 0.7450980392156863, 0.8117647058823529}

\newcommand{\dark}[1]{#1!50!black}
\newcommand{\light}[1]{#1!50!white}

\tikzset{color/.style={draw=\dark{#1}, fill=\light{#1}}}

\tikzset{block/.style={rectangle, minimum width=3.75ex, minimum height=3.75ex, inner sep=-0.5ex}}

\colorlet{color1}{mplblue!50!white}
\colorlet{color2}{mplorange!50!white}
\colorlet{color3}{mplred!50!white}
\colorlet{color4}{mplpurple!50!white}
\colorlet{color5}{mplgreen!50!white}
\colorlet{color6}{mplgray!50!white}

\colorlet{gray1}{gray!90!black}
\colorlet{gray2}{gray!60!white}
\colorlet{gray3}{gray!30!white}

\tikzset{line/.style={draw=gray1, semithick}}
\tikzset{arrow/.style={-stealth, line}}

\usepackage[framemethod=TikZ]{mdframed}

\begin{document}

 \cfoot{\thepage}
 \setlength{\footskip}{3em}

\twocolumn[
\icmltitle{%
A Simple Zero-shot Prompt Weighting Technique to Improve Prompt Ensembling in Text-Image Models}

\icmlsetsymbol{equal}{*}
\icmlsetsymbol{wdagr}{$\dagger$}

\begin{icmlauthorlist}
\icmlauthor{James Urquhart Allingham}{equal,wdagr,cam}
\icmlauthor{Jie Ren}{equal,dm}
\icmlauthor{Michael W. Dusenberry}{dm} \\
\icmlauthor{Xiuye Gu}{gr}
\icmlauthor{Yin Cui}{wdagr,nv}
\icmlauthor{Dustin Tran}{dm}
\icmlauthor{Jeremiah Zhe Liu}{wdagr,gr,har}
\icmlauthor{Balaji Lakshminarayanan}{dm}
\end{icmlauthorlist}

\icmlaffiliation{cam}{University of Cambridge}
\icmlaffiliation{dm}{Google DeepMind}
\icmlaffiliation{gr}{Google Research}
\icmlaffiliation{nv}{Nvidia}
\icmlaffiliation{har}{Harvard University}

\icmlcorrespondingauthor{James Urquhart Allingham}{jua23@cam.ac.uk}
\icmlcorrespondingauthor{Jie Ren}{jjren@google.com}

\icmlkeywords{Machine Learning, ICML}

\vskip 0.3in
]

\printAffiliationsAndNotice{\icmlEqualContribution} %

\begin{abstract}
Contrastively trained text-image models have the remarkable ability to perform zero-shot classification, that is, classifying previously unseen images into categories that the model has never been explicitly trained to identify. However, these zero-shot classifiers need prompt engineering to achieve high accuracy.  Prompt engineering typically requires hand-crafting a set of prompts for individual downstream tasks. 
In this work, we aim to automate this prompt engineering and improve zero-shot accuracy through prompt ensembling. In particular, we ask \emph{``Given a large pool of prompts, can we automatically score the prompts and ensemble those that are most suitable for a particular downstream dataset, without needing access to labeled validation data?"}. We demonstrate that this is possible. In doing so, we identify several pathologies in a naive prompt scoring method where the score can be easily overconfident due to biases in pre-training and test data, and we propose a novel prompt scoring method that corrects for the biases.
Using our proposed scoring method to create a weighted average prompt ensemble, our method outperforms an equal average ensemble, as well as hand-crafted prompts, 
on ImageNet, 4 of its variants, and 11 fine-grained classification benchmarks, all while being  fully automatic, optimization-free, and not requiring access to labeled validation data.%
\end{abstract}
\vspace{-2em}

\section{Introduction}
\label{sec:intro}

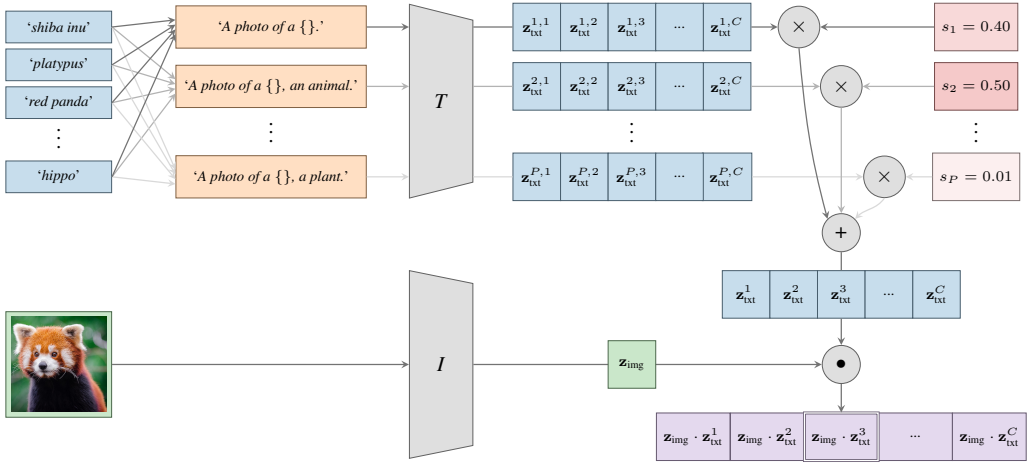
\begin{figure*}[ht]
    \centering
    \resizebox{0.8\linewidth}{!}{
    \begin{tikzpicture}
    \tikzset{class/.style={color=color1, minimum width=5em, minimum height=1.5em, font=\scriptsize}}
    
    \node[class] at (0,0) (t1) {`\emph{shiba inu}'};
    \node[class, below=0.25 em of t1] (t2) {`\emph{platypus}'};
    \node[class, below=0.25 em of t2] (t3) {`\emph{red panda}'};
    \node[below=-0.5 em of t3] (t4) {\vdots};
    \node[class, below=0.25 em of t4] (t5) {`\emph{hippo}'};

    \tikzset{prompt/.style={color=color2, minimum width=9em, minimum height=2.em, font=\scriptsize}}
    
    \node[prompt, right = 3.em of t1.east] (p1) {`\emph{A photo of a \{\}.}'};
    \node[prompt] at (t3.north east -| p1) (p2) {`\emph{A photo of a \{\}, an animal.}'};
    \node[below=-0.25 em of p2] (dots) {\vdots};
    \node[prompt] at (t5.east -| p1) (p3) {`\emph{A photo of a \{\}, a plant.}'};

    \foreach \src/\ofset in {t1/0.75ex, t2/0.25ex, t3/-0.25ex, t5/-0.75ex} {
        \draw[arrow, gray2] (\src.east) -- ([yshift=\ofset]p2.west);
    }
    
    \foreach \src/\ofset in {t1/0.75ex, t2/0.25ex, t3/-0.25ex, t5/-0.75ex} {
        \draw[arrow, gray3] (\src.east) -- ([yshift=\ofset]p3.west);
    }
    
    \foreach \src/\ofset in {t1/0.75ex, t2/0.25ex, t3/-0.25ex, t5/-0.75ex} {
        \draw[arrow] (\src.east) -- ([yshift=\ofset]p1.west);
    }

    \node[draw, trapezium, color=color6, rotate=-90, trapezium stretches body, text width=7em, minimum height=3em, align=center, above right= -0.2em and 2.em of p2] (T) {\rotatebox{90}{\emph{T}}};
    
    \draw[arrow] (p1.east) -- (p1.east -| T.south);
    \draw[arrow, gray2] (p2.east) -- (p2.east -| T.south);
    \draw[arrow, gray3] (p3.east) -- (p3.east -| T.south);

    \node[coordinate, right=3.em of T.north] (anchor) {}; 
    \foreach \row/\i in {p1/1, p2/2, p3/P} {
        \foreach \ofset/\j in {0em/1, 2.25em/2, 4.5em/3, 9em/C}
            \node[color=color1, xshift=\ofset, minimum width=2.25em, minimum height=2.25em] at (\row -| anchor) (z\i\j) {\scriptsize $\mathbf{z}_\text{txt}^{\i,\j}$};
    }
    \foreach \row in {p1, p2, p3} {
        \node[color=color1, xshift=6.75em, minimum width=2.25em, minimum height=2.25em] at (\row -| anchor) {\scriptsize ...};
    }
    
    \draw[line] (p1.east -| T.north) -- (z11);
    \draw[line, gray2] (p2.east -| T.north) -- (z21);
    \draw[line, gray3] (p3.east -| T.north) -- (zP1);
    
    \node[] at (dots -| z13){\vdots};

    \tikzset{opnode/.style={circle, minimum width=4ex, color=color6}}
    \tikzset{mul/.style={opnode, font={$\times$}}}

    \node[mul, right = 1.25em of z1C.east] (m1) {};
    \node[mul, right = 3.25em of z2C.east] (m2) {};
    \node[mul, right = 5.25em of zPC.east] (m3) {};
    
    \draw[arrow] (z1C.east) -- (m1);
    \draw[arrow, gray2] (z2C.east) -- (m2);
    \draw[arrow, gray3] (zPC.east) -- (m3);
    
    \node[color=color3!30!white, minimum width=2.25em, minimum height=2.25em, right=1.25em of m3.east] (s3) {\scriptsize $s_P = 0.01$};
    \node[color=color3, minimum width=2.25em, minimum height=2.25em] at (m2 -| s3) (s2) {\scriptsize $s_2 = 0.50$};
    \node[color=color3!70!white, minimum width=2.25em, minimum height=2.25em] at (m1 -| s3) (s1) {\scriptsize $s_1 = 0.40$};
    
    \draw[arrow] (s1) -- (m1.east);
    \draw[arrow, gray2] (s2) -- (m2.east);
    \draw[arrow, gray3] (s3) -- (m3.east);
    
    \node[] at (dots -| s3) {\vdots};

    \node[color=color6, below = 5.em of m2.south, opnode] (add) {+};
    
    \draw[arrow] (m1.south) to[out=-85,in=105] (add.north west);
    \draw[arrow, gray2] (m2.south) -- (add.north);
    \draw[arrow, gray3] (m3.south) to[out=-125,in=45] (add.north east);

    \node[coordinate, below=2.em of add.south] (anchor2) {}; 
    \foreach \ofset/\j in {-4.5em/1, -2.25em/2, 0em/3, 4.5em/C} {
        \node[color=color1, xshift=\ofset, minimum width=2.25em, minimum height=2.25em] at (anchor2) (z\j) {\scriptsize $\mathbf{z}_\text{txt}^{\j}$};
    }
    \node[color=color1, xshift=2.25em, minimum width=2.25em, minimum height=2.25em] at (anchor2) {\scriptsize ...};
    
    \draw[line] (add.south) -- (z3.north);

    \tikzset{dot/.style={opnode, font={$\bullet$}}}
    \node[dot, below = 1.25em of z3.south] (dot) {};
    
    \draw[arrow] (z3.south) -- (dot.north);

    \node[draw, trapezium, color=color6, rotate=-90, trapezium stretches body, text width=7em, minimum height=3em, align=center] at (dot -| T) (I) {\rotatebox{90}{\emph{I}}};
    
    \draw[arrow] (I.north) -- (dot.west);

    \node[color=color5, minimum width=2.25em, minimum height=2.25em] at (dot -| zP3) (zimg) {\scriptsize $\mathbf{z}_\text{img}$};

    \node[inner sep=0.25em, color=color5] (red) at (I -| t3)
    {\includegraphics[width=4.5em]{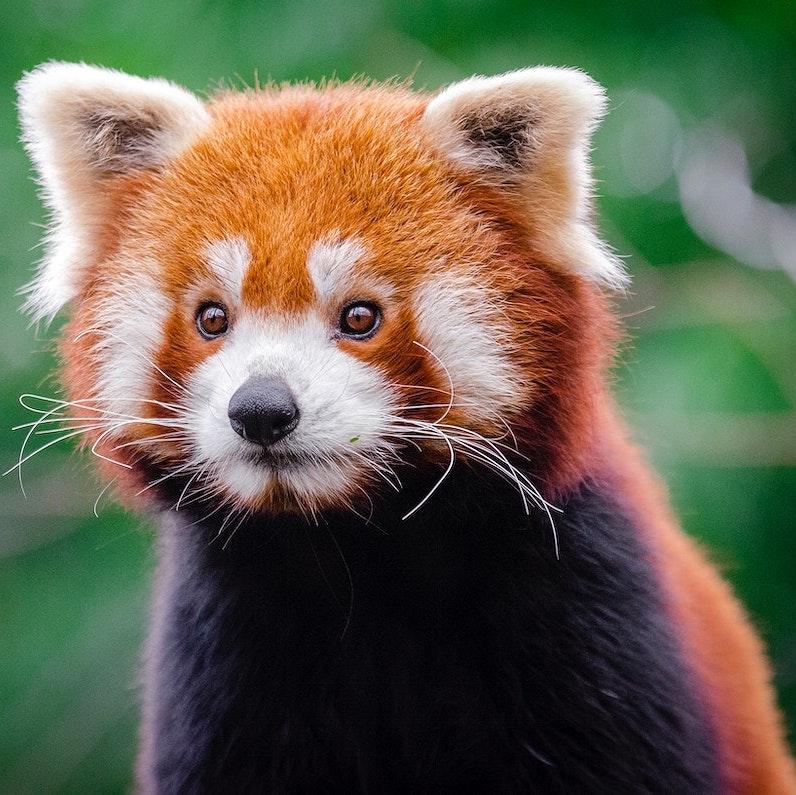}};
    
    \draw[arrow] (red.east) -- (I.south);

    \node[coordinate, below=2.5em of dot.south] (anchor3) {}; 
    \foreach \ofset/\j in {-7em/1, -3.5em/2, 7em/C} {
        \node[color=color4, xshift=\ofset, minimum width=2.25em, minimum height=2.25em] at (anchor3) (l\j) {\scriptsize $\mathbf{z}_\text{img} \cdot \mathbf{z}_\text{txt}^{\j}$};
    }
    
    \node[color=color4, xshift=3.5em, minimum width=3.5em, minimum height=2.25em] at (anchor3) {\scriptsize ...};
    
    \node[color=color4, double, xshift=0, minimum width=2.25em, minimum height=2.25em] at (anchor3) (l3) {\scriptsize $\mathbf{z}_\text{img} \cdot \mathbf{z}_\text{txt}^{3}$};
    
    \draw[arrow] (dot.south) -- (l3.north);

    \end{tikzpicture}
    }
    \tikzset{smallpatch/.style={rectangle, minimum width=1.0ex, minimum height=1.0ex}}
    \newcommand{\patchmarker}[1][mplgray]{\scalerel*{\tikz \node[smallpatch, color=#1] {};}{\square}}
    \vspace{-1em}
    \caption{Construction of a zero-shot classifier with zero-shot prompt ensembling (ZPE) for text-image models. Logits (\patchmarker[color4]) are calculated by combining text (\patchmarker[color1]) and image (\patchmarker[color5]) representations. The final text representation is a weighted ensemble of representations corresponding to different prompts (\patchmarker[color2]). Crucially, the ZPE scores (\patchmarker[color3]) for weighting each prompt are calculated without access to any labeled training data, as described in \cref{sec:zpe,alg:normalised_max_logit}.}
    \label{fig:zpe_classifier_setup}
    \vspace{-1.25em}
\end{figure*}

Contrastively trained text-image models such as CLIP \citep{radford2021learning}, ALIGN \citep{jia2021scaling}, LiT \citep{zhai2021lit}, and BASIC \citep{pham2021combined} have the remarkable ability to perform \emph{zero-shot}\footnote{Here, zero-shot refers to the fact that the classifier has not been trained in a supervised manner using any examples of the class. However, due to the large-scale pre-training of the models, it is possible that relevant examples were observed. For this reason, this setting is often referred to as zero-shot \emph{transfer} rather than zero-shot \emph{learning} \citep{zhai2021lit}.} classification. 
That is, such models can be used to classify previously unseen images into categories for which the model has never been explicitly trained to identify. 
Such zero-shot classifiers can match the performance of standard classification models which have access to training examples.
For example, a CLIP zero-shot classifier with a ViT-L/14 vision tower matches the ImageNet accuracy of a ResNet-101 \citep{radford2021learning}.
However, achieving strong zero-shot classification performance requires prompt engineering \citep{radford2021learning, zhai2021lit, pham2021combined}.
Zero-shot CLIP ViT-B/16 performance on ImageNet increases from 64.18\% to 66.92\% and 68.57\% when using the prompt `\emph{A photo of \{\}.}', and a selection of 80 hand-crafted prompts, rather than class name only.
To use a set of hand-crafted prompts for zero-shot classification, the text embeddings of the prompts composed with class names are averaged into a single vector to represent the class. This is called a \textit{prompt ensemble} in \cite{radford2021learning}.
Prompt engineering can be seen as reducing the `distribution shift' between the zero-shot setting and the training data in which the captions seldom consist of a single word.

Unfortunately, the need for a set of hand-crafted prompts to achieve good zero-shot performance greatly reduces the promised general applicability of such zero-shot classifiers. 
Different sets of prompts were manually designed and tuned for different downstream tasks for CLIP.
For example, the prompts `\emph{a photo of a \{\} texture.}', `\emph{a photo of a \{\} pattern.}', and `\emph{i love my \{\}!}', `\emph{a photo of my clean \{\}.}' were designed for the Describable Textures Dataset (DTD) 
\citep{cimpoi2014describing} 
and the Cars196 dataset 
\citep{krause20133d}
, respectively.
Designing different sets of hand-crafted prompts can be be labor-intensive, and 
the common prompt design processes require access to a labeled validation dataset, which may not be available in practice. 
In this paper, we ask the question \emph{``Can we automate prompt engineering for zero-shot classifiers?''}. 
Specifically, given a zero-shot model and a large pool of potential prompts, our goal is to select the optimal subset of prompts that maximize the model performance in a \textit{zero-shot} fashion, i.e., without access to a labeled validation set. 

Our contributions are the following:
\begin{enumerate}[itemsep=0ex]
    \item We present an algorithm for automatically scoring the importance of prompts in a large pool given a specific downstream task when using text-image models for zero-shot classification. We then propose a weighted average prompt ensembling method using the scores as the weights. 
    
    \item We identify several pathologies in a naive prompt scoring method where the score can be easily overconfident due to biases in both pre-training and test data. We address these pathologies via bias correction in a zero-shot and optimization-free fashion.
    
    \item We demonstrate that our algorithm is better performing than the existing approach of hand-crafted prompts, without the need for a labeled validation set and a labor-intensive manual tuning process.
    
\end{enumerate}

\section{Background}

We consider contrastively trained text-image models consisting of a text encoder $T$ and an image encoder $I$. 
The encoders produce embeddings $\mathbf{z}_\text{txt} = T(\mathtt{text})$ and $\mathbf{z}_\text{img} = I(\mathtt{image})$, both of size $D$. 

\paragraph{Training.} The models are trained on batches of $B$ text-image pairs $\{(\mathtt{text}_b, \mathtt{image}_b)\}_{b=1}^B$---e.g., photographs and their captions---to encourage that $\mathbf{z}_\text{txt}^i = \mathbf{z}_\text{img}^j$, if $i = j$, and $\mathbf{z}_\text{txt}^i \ne \mathbf{z}_\text{img}^j$ otherwise.
This is accomplished with a \emph{bi-directional contrastive loss}:
\allowdisplaybreaks[1]
{
\begin{align*}
    \mathtt{logits} &= \begin{bmatrix}
        \mathbf{z}_\text{img}^1 \cdot \mathbf{z}_\text{txt}^1 &\mathbf{z}_\text{img}^1 \cdot \mathbf{z}_\text{txt}^2 & \dots  & \mathbf{z}_\text{img}^1 \cdot \mathbf{z}_\text{txt}^B \\
        \mathbf{z}_\text{img}^2 \cdot \mathbf{z}_\text{txt}^1 &\mathbf{z}_\text{img}^2 \cdot \mathbf{z}_\text{txt}^2 & \dots  & \mathbf{z}_\text{img}^2 \cdot \mathbf{z}_\text{txt}^B \\
        \vdots & \vdots & \ddots & \vdots \\
        \mathbf{z}_\text{img}^B \cdot \mathbf{z}_\text{txt}^1 &\mathbf{z}_\text{img}^B \cdot \mathbf{z}_\text{txt}^2 & \dots  & \mathbf{z}_\text{img}^B \cdot \mathbf{z}_\text{txt}^B 
    \end{bmatrix} \\
    \mathcal{L}_\text{img} &= -\frac{1}{B} \sum_{b=1}^B \log \mathrm{softmax}(\mathtt{logits})_{b,b}\\
    \mathcal{L}_\text{txt} &= -\frac{1}{B} \sum_{b=1}^B \log \mathrm{softmax}(\mathtt{logits}^\intercal)_{b,b}\\
    \mathcal{L} &= \left(\mathcal{L}_\text{img} + \mathcal{L}_\text{txt}\right)/2, 
\end{align*}
}%
which can be interpreted as the average cross-entropy loss when classifying which caption in the batch corresponds to a given image and vice-versa.

\paragraph{Zero-shot prediction.} Once the text and image encoders have been trained, we can setup a zero-shot classifier with $C$ classes for an image with representation $\mathbf{z}_\text{img}$ by computing
\begin{align*}
    \mathtt{logits} &= \begin{bmatrix} \mathbf{z}_\text{img} \cdot \mathbf{z}_\text{txt}^1 & \dots  &\mathbf{z}_\text{img} \cdot \mathbf{z}_\text{txt}^c & \dots & \mathbf{z}_\text{img} \cdot \mathbf{z}_\text{txt}^C \end{bmatrix} \\
    \hat{c} &= \mathrm{argmax}_c\ \mathtt{logits},
\end{align*}
where 
$\hat{c}$ is the predicted class, and 
$\mathbf{z}_\text{txt}^c = T(\mathtt{class\_name}_c)$ with $\mathtt{class\_name}$ being a list of possible classes.
In the case where we have $P$ prompt templates, prompt ensembling as proposed in \citep{radford2021learning} generalizes the above to
\begin{align}
    \hspace{-0.65em}\mathtt{logits} \hspace{-0.2em}&=\hspace{-0.2em} \begin{bmatrix}
        \mathbf{z}_\text{img} \cdot \mathbf{z}_\text{txt}^{1,1} &\mathbf{z}_\text{img} \cdot \mathbf{z}_\text{txt}^{1,2} & \dots  & \mathbf{z}_\text{img} \cdot \mathbf{z}_\text{txt}^{1,C} \\
        \mathbf{z}_\text{img} \cdot \mathbf{z}_\text{txt}^{2,1} &\mathbf{z}_\text{img} \cdot \mathbf{z}_\text{txt}^{2,2} & \dots  & \mathbf{z}_\text{img} \cdot \mathbf{z}_\text{txt}^{2,C} \\
        \vdots & \vdots & \ddots & \vdots \\
        \mathbf{z}_\text{img} \cdot \mathbf{z}_\text{txt}^{P,1} &\mathbf{z}_\text{img} \cdot \mathbf{z}_\text{txt}^{P,2} & \dots  & \mathbf{z}_\text{img} \cdot \mathbf{z}_\text{txt}^{P,C}
    \end{bmatrix}\hspace{-0.4em} \\
    \hat{c} &= \mathrm{argmax}_c\ \frac{1}{P} \sum_{p = 1}^P \mathtt{logits}_{p}\, , \label{eq:prompt_ensemble}
\end{align}
where $\mathtt{logits}_{p}$ is the $p^\mathrm{th}$ row of $\mathtt{logits}$, and $\mathbf{z}_\text{txt}^{p,c} = T(\mathtt{prompt\_template}_p \circ \mathtt{class\_name}_c)$, with $\circ$ indicating the composition of a prompt template and a class name, e.g., `\emph{A photo of a $\{\}$.}' $\circ$ `\emph{dog}' $=$ `\emph{A photo of a dog.}'. 
Note that \cref{eq:prompt_ensemble} can be seen as constructing an ensemble of classifiers in logit space, where each classifier uses a different prompt.

\section{Zero-shot weighted prompt ensembling} \label{sec:zpe}

In this section, we describe our proposed method: Zero-shot Prompt Ensembling (ZPE) for zero-shot classification with text-image models.
Given a large pool of $P$ prompts, which may or may not be entirely relevant to a specific problem at hand, and a previously unseen classification task, we would like to learn a set of scores $\{s_1, s_2, \dots, s_P\}$ that will allow us to perform a weighted average by replacing \cref{eq:prompt_ensemble} with 
\begin{align}
    \hat{c} &= \mathrm{argmax}\ \frac{1}{P} \sum_{p = 1}^P \mathtt{logits}_{p} \mathbin{\color{Green} \times} \textcolor{Green}{s_p}, \label{eq:weighted_prompt_ensemble}
\end{align}
or a masked average by replacing \cref{eq:prompt_ensemble} with 
\begin{align}
    \hat{c} &= \mathrm{argmax}\ \frac{1}{P} \sum_{p = 1}^P \mathtt{logits}_{p} \mathbin{\color{Green} \times} \textcolor{Green}{s_p \times \mathds{1}(s_p > \tau)}, \label{eq:masked_prompt_ensemble}
\end{align}
where $\mathds{1}(\cdot)$ is the indicator function.
The masked average introduces a hyperparameter $\tau$, the score threshold for prompt subset selection.
Thus, we also consider the hyperparameter-free weighted average.
The weighted averaging of logits can be regarded as the weighted ensemble of many classifiers, where each of them is made of a different prompt, and the weights are computed in a zero-shot fashion without any optimization or the access to test labels.

\subsection{A simple baseline -- max logit scoring}

The maximum logit over the classes $\mathtt{max\_logit} = \mathrm{max}_c\ \mathtt{logits}$ is a commonly used confidence score for classification problems. 
Since $\mathbf{z}_\text{img}$ and $\mathbf{z}_\text{txt}$ are $L_2$ normalized, i.e. $\|\mathbf{z}_\text{img}\|=\|\mathbf{z}_\text{txt}\|=1$, the inner product equals to the $L_2$ distance up to a constant and a scalar, i.e.,  $\|\mathbf{z}_\text{img} - \mathbf{z}_\text{txt}\|^2 = 2 -2*\mathbf{z}_\text{img} \cdot {\mathbf{z}_\text{txt}}^\intercal$. Thus, the maximum logit over the classes is equivalent to the minimum $L_2$ distance over the classes. 
The minimum distance is a natural measure of confidence for a prediction. For example, the classic $k$-means algorithm uses it as a measure for clustering \citep{macqueen1967classification}. 
Recent work also has shown that the maximum logit outperforms the maximum softmax in terms of capturing the uncertainty in prediction \citep{hendrycks2019scaling}, and it was used as a confidence score in zero-shot classification with text-image models \citep{ge2022improving}. 

For the problem of prompt scoring, intuitively, if a prompt has large maximum logit values given a set of images, it suggests that the zero-shot classifier is more confident in the predictions, and therefore it is more likely that the prompt is suitable for the image classification task. 
Thus, we consider \cref{alg:max_logit} for using the maximum logit (averaged over images) for scoring prompts.

\begin{algorithm}[h]
   \caption{\textsc{Max Logit Scoring}}
   \label{alg:max_logit}
\begin{algorithmic}[1]
   \STATE {\bfseries Input:} Image embeddings $\mathbf{Z}_\text{img}$ (shape $N \times D$), class embeddings for the $p^\text{th}$ prompt $\mathbf{Z}_\text{txt}$ (shape $C \times D$).
   \STATE $\mathtt{logits} = \mathbf{Z}_\text{img} \cdot {\mathbf{Z}_\text{txt}}^\intercal$ \hfill \# shape: $N \times C$.
   \STATE $\mathtt{max\_logits} = \mathrm{max}_c\ \mathtt{logits}$ \hfill \# shape: $N$.
   \STATE $s_p = \frac{1}{N}\sum_{n=1}^N\mathtt{max\_logits}_n$
   \STATE {\bfseries Output:} $s_p$, the score for the $p^\text{th}$ prompt.
\end{algorithmic}
\end{algorithm}

\begin{figure}[t]
    \centering
    \begin{subfigure}[b]{0.325\linewidth}
        \centering
        \includegraphics[width=\linewidth]{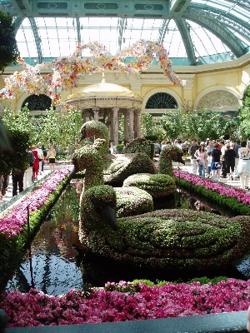}
        \caption{Topiary}
        \label{fig:first}
    \end{subfigure}
    \hfill
    \begin{subfigure}[b]{0.325\linewidth}
        \centering
        \includegraphics[width=\linewidth]{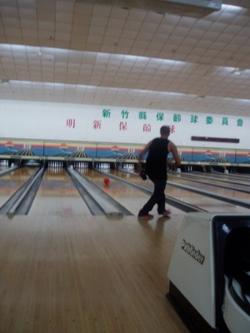}
        \caption{Bowling alley}
        \label{fig:second}
    \end{subfigure}
    \hfill
    \begin{subfigure}[b]{0.325\linewidth}
        \centering
        \includegraphics[width=\linewidth]{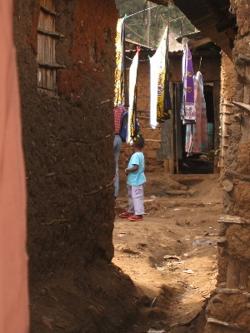}
        \caption{Medina}
        \label{fig:third}
    \end{subfigure}
    \vspace{-2.em}
    \caption{Images from Sun397. Note that the pictures contain the spurious `person' concept which is different from the classes.}
    \vspace{-1.5em}
    \label{fig:sun_imgs}
\end{figure}

Unfortunately, while this scoring method does work to some extent, it is biased.
The biases can easily be seen by looking at the top 10 prompts for ImageNet and Sun397:
\begin{tcolorbox}[colback=mplred!3,colframe=mplred!75!white,title=\textsc{ImageNet},left=0.5ex,right=0.5ex,top=0.5ex,bottom=0.5ex]
\fontsize{9}{9}\selectfont
\emph{a example of a \textbf{person} practicing \{\}.}$\,\cdot\,$
\emph{a example of a \textbf{person} using \{\}.}$\,\cdot\,$
\emph{a cropped photo of a \{\}.}$\,\cdot\,$
\emph{a photo of the \{\}.}$\,\cdot\,$
\emph{a photo of the small \{\}.}$\,\cdot\,$
\emph{a cropped photo of the \{\}.}$\,\cdot\,$
\emph{a photo of the large \{\}.}$\,\cdot\,$
\emph{a example of the \textbf{person} \{\}.}$\,\cdot\,$
\emph{a example of a \textbf{person} \{\}.}$\,\cdot\,$
\emph{a example of \{\}.}
\end{tcolorbox}
\begin{tcolorbox}[colback=mplred!3,colframe=mplred!75!white,title=\textsc{Sun397},left=0.5ex,right=0.5ex,top=0.5ex,bottom=0.5ex]
\fontsize{9}{9}\selectfont
\emph{a photo of \{\}.}$\,\cdot\,$
\emph{a photo of the \{\}.}$\,\cdot\,$
\emph{a cropped photo of a \{\}.}$\,\cdot\,$
\emph{a example of \{\}.}$\,\cdot\,$
\emph{a example of the \textbf{person} \{\}.}$\,\cdot\,$
\emph{an example of \{\}.}$\,\cdot\,$
\emph{a example of a \textbf{person} \{\}.}$\,\cdot\,$
\emph{a photo of a large \{\}.}$\,\cdot\,$
\emph{a photo of the large \{\}.}$\,\cdot\,$
\emph{a cropped photo of the \{\}.}
\end{tcolorbox}
We see that some prompts---i.e., those containing the word `person'---are scored highly even though the these prompts are not related to the classes of either dataset.
By considering the contrastive training of our model we can identify two pathologies that might cause this problem.
Prompts are biased towards large logits due to 
\begin{itemize}
    \item \textbf{Word frequency bias in pre-training data:} Prompts containing words, or words with similar semantic meanings to those, that appear more frequently in the pre-training data\footnote{
        E.g., `women', `men', `baby', `kids', `man', `girl', and `woman', which are all semantically similar to `person', are included in the top 100 most frequent words of LAION400m.
    }, and
    \item \textbf{Spurious concept frequency bias in test data:} Prompts containing frequent words that map to common concepts in the test images, but that are different to the classes of interest for prediction. For example, images in Sun397 often contain people but the classes are various in- and outdoor locations; see \cref{fig:sun_imgs}.
\end{itemize}

This suggests that the raw max logit score is not trustworthy because the value can be overconfident due to the biases.

\subsection{Tackling frequency biases via logit normalization}

To correct for these frequency biases, we consider normalizing the raw max logits score by subtracting the expected value under a reference distribution. 
We use subtraction rather than division for normalization because we are working in log (odds) space.
This approach is inspired by the classical likelihood ratio method which compares how different the likelihood evaluated at the observed data is from that evaluated at a reference data \cite{casella2021statistical,king1989unifying}, and it is a commonly used technique for reducing biases.
For example, the Term Frequency–Inverse Document Frequency (TF-IDF) in information retrieval \citep{jones1972statistical}, the likelihood-ratio method and relative Mahalanobis distance method for out-of-distribution detection \citep{ren2019likelihood,ren2021simple,ren2022out}, and the ratio of observed to expected mortality rate (O/E) in medical studies \citep{best1994ratio,galvan2015observed} all use the expectation as a reference to normalize the raw scores. See Section \ref{sec:related_work} for further discussion. 

Given a pair of a test image and a prompt, we compare the maximum logit for the pair $\mathtt{logits}(\mathtt{test\_img}, \mathtt{prompt})$ with the expected maximum logit for a random image with the same prompt $\mathtt{logits}(\mathtt{random\_img}, \mathtt{prompt})$.  
If the prompt contains frequent words in the pre-training data or words that map to unrelated but common concepts in the test data and result in large logits regardless of the content of an image, the expected maximum logit value %
would be large too. 
Therefore only when the difference $\mathtt{logits}(\mathtt{img}, \mathtt{prompt}) - \mathtt{logits}(\mathtt{random\_img}, \mathtt{prompt})$ is large, the prompt is considered suitable to the classification task of the test image. 

We solve \emph{word frequency bias} by normalizing the logits for each prompt by subtracting
the expected logits based on images in the pre-training data $E_{\text{pretrain}} = \mathbb{E}_{\mathtt{img}\sim \mathcal{D}_{\text{pretrain}}}\left[\mathbf{z}_\text{img} \cdot \mathbf{z}_\text{txt}\right]$.
We estimate the expected logits using the average logits for a wide range of random images sampled from the pre-training data. 
Since the pre-training data of CLIP is not publicly available, we instead use LAION400m \citep{schuhmann2021laion}, since it has been shown that the models pre-trained using LAION dataset could reproduce CLIP models' performance \citep{cherti2022reproducible}. 
By removing $E_{\text{pre-train}}$, we down-weigh prompts that contain the frequent words in the pre-training data which would result in large logits regardless of the content of an image.
In the experiments, we use a small subsample of LAION400m, i.e., the first 20k images, as we found it is already sufficient to achieve high performance.

As a sanity check, we verify that subtracting $E_{\text{pretrain}}$ from the logits reduces the word frequency bias. We compare the correlation coefficient for the frequency of each word in LAION400m and the average logit $\mathtt{avg\_logit} = \mathbb{E}_{\mathtt{img}\sim \mathcal{D}_{\text{ImagNet}}}\left[\mathbf{z}_\text{img} \cdot \mathbf{z}_\text{word}\right]$ with the correlation between the word frequency and $\mathtt{avg\_logit} - E_{\text{pretrain}}$. Without normalisation we have a Pearson correlation coefficient of 0.09 with a p-value of $3.6\times10^{-11}$. With normalisation we have a correlation coefficient of -0.03 with a p-value of 0.03. That is, subtracting $E_{\text{pretrain}}$ removes the statistically significant correlation between logit magnitude and word frequency.

We solve \emph{spurious concept frequency bias} subtracting the expected logits for the images in the test data itself $E_{\text{test}} = \mathbb{E}_{\mathtt{img}\sim \mathcal{D}_{\text{test}}}\left[\mathbf{z}_\text{img} \cdot \mathbf{z}_\text{txt}\right]$. 
Our intuition is that if there are spurious but common concepts shared among test images, averaging the logits would provide a good reference of the maximum logit value for an general image containing the concepts. 
By removing $E_{\text{test}}$, we down-weigh prompts that contain words that map to common but spurious concepts. 

To jointly reduce the both types of biases, we average over $E_{\text{pretrain}}$ and $E_{\text{test}}$. In summary, \cref{alg:normalised_max_logit} shows our method for scoring prompts with normalization.

\begin{algorithm}[h]
   \caption{\textsc{Normalized Max Logit Scoring}}
   \label{alg:normalised_max_logit}
\begin{algorithmic}[1]
   \STATE {\bfseries Input:} Image embeddings $\mathbf{Z}_\text{img}$ (shape $N \times D$), class embeddings for the $p^\text{th}$ prompt $\mathbf{Z}_\text{txt}$ (shape $C \times D$), \textcolor{Green}{embeddings of pre-train images $\mathbf{Z}_\text{pretrain}$ (shape $N' \times D$).}
   \STATE $\mathtt{logits} = \mathbf{Z}_\text{img} \cdot {\mathbf{Z}_\text{txt}}^\intercal$ \hfill \# shape: $N \times C$.
   \STATE \textcolor{Green}{$\mathtt{logits_\text{pretrain}} = \mathbf{Z}_\text{pretrain} \cdot {\mathbf{Z}_\text{txt}}^\intercal$ \hfill \# shape: $N' \times C$.}
   \STATE \textcolor{Green}{$E_{\text{pretrain}} =  \frac{1}{N'}\sum_{n=1}^{N'}\mathtt{logits_\text{pretrain,n}}$}
   \STATE \textcolor{Green}{$E_{\text{test}} =  \frac{1}{N}\sum_{n=1}^{N}\mathtt{logits}_n$}
   \STATE \textcolor{Green}{$\mathtt{logits_\text{normalized}} = \mathtt{logits} - (E_{\text{pretrain}} +  E_{\text{test}}) / 2$}
   \STATE $\mathtt{max\_logits} = \mathrm{max}_c\ \mathtt{logits}_{\textcolor{Green}{\text{normalized}}}$ \hfill \# shape: $N$.
   \STATE $s_p = \frac{1}{N}\sum_{n=1}^N\mathtt{max\_logits}_n$
   \STATE {\bfseries Output:} $s_p$, the score for the $p^\text{th}$ prompt.
\end{algorithmic}

\end{algorithm}
\vspace{-0.5em}

\subsection{Handling long-tails via softmax weighting}

\begin{figure}
    \centering
    \includegraphics[width=\linewidth]{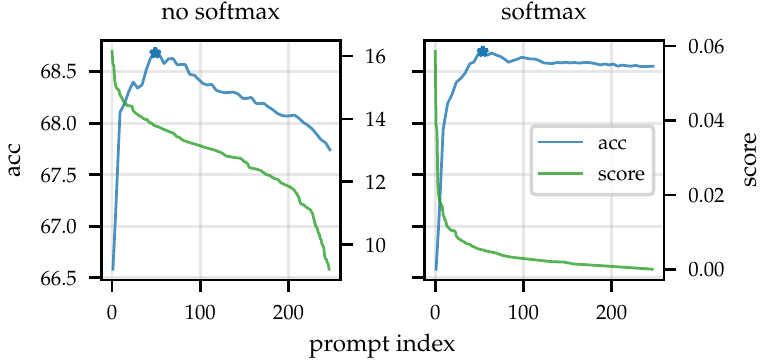}
    \vspace{-1.5em}
    \caption{CLIP ViT-B/16 ImageNet accuracy (given a weighted ensemble of all prompts up to the prompt index on the x-axis) and the ZPE scores corresponding to each prompt, with and without softmax weighting.}
    \label{fig:acc_score}
\end{figure}

When scoring a large number of prompts, we observe a long tail behaviour, where a small number of prompts have large scores, but most prompts are ``bad'' and have  small scores. 
Despite receiving small scores, the irrelevant prompts can still collectively provide a large impact on the weighted average in \cref{eq:weighted_prompt_ensemble}.
To mitigate this issue, we replace \cref{eq:weighted_prompt_ensemble} with
\begin{align}
    \vspace{-0.75em}\mathtt{c} \vspace{-0.2em}&= \vspace{-0.2em} \mathrm{argmax}\ \!\frac{1}{P}\! \sum_{p = 1}^P \mathtt{logits}_{p} \times \textcolor{Green}{\mathrm{softmax}(} s\textcolor{Green}{)}_p.\vspace{-0.4em} \label{eq:softmax_prompt_ensemble}
\end{align}
The softmax function is applied over the prompt sets such that the weights are summed to 1.
\cref{fig:acc_score} shows the impact of (i) the long tail of low ZPE scores on weighted prompt ensemble ImageNet accuracy, and (ii) the reduced impact when using softmax weighting.
Note that without softmax weighting, adding additional prompts has a large negative effect on accuracy once irrelevant prompts start to be included in the ensemble.

\subsection{Prompt selection}

So far we have considered principled normalization techniques for countering the frequency biases in pre-training and test data and the long-tail issue in computing prompt scores $s_p$ for the weighted ensemble (\ref{eq:weighted_prompt_ensemble}). For computing the masked ensemble (\ref{eq:masked_prompt_ensemble}), we also need to set the score threshold parameter $\tau$. Here, we consider prompt selection as an outlier detection problem. We assume that in a large pool of prompts, the majority of the prompts are irrelevant to a specific downstream dataset. Thus, the relevant prompts can be regarded as outliers to the pool distribution. 
We use the median absolute deviation \citep{rousseeuw1993alternatives} test statistic which is similar to using a $z$-test statistic, but is more robust to extreme events and non-Gaussian distributions.
Concretely, we calculate the median $\bar{s} = \mathrm{median}_p\left(s \right)$ and median absolute deviation $\tilde{d} = \mathrm{median}_p \left(|s - \bar{s}|\right)$. We then compute the $z$-score for given a prompt, 
\begin{align*}
    z &= \frac{s - \bar{s}}{\tilde{d}}.
\end{align*}
We classify the $p^\text{th}$ prompt as an outlier if $z_p > \tau$.
Here, $\tau$ is analogous to a desired standard deviation in a $z$-test.
The advantage of this approach, rather than thresholding the scores directly, is that we can set $\tau$ without knowledge of the magnitudes of the scores.  
This allows us to use the same value of $\tau$ for multiple datasets.

\section{Experimental Evaluation}

To evaluate the quality of our prompt scores,
we compare our zero-shot prompt ensembling (ZPE) method, to a number of baselines. 
We also perform various ablation and sensitivity studies.
Unless otherwise specified, we use %
score normalization, and softmax weighting.
We evaluate the methods on ImageNet \citep{russakovsky2015imagenet}, and its variant test sets ImageNet-R, ImageNet-A, ImageNet-Sketch, and ImageNet-V2 \citep{hendrycks2021many,hendrycks2021natural,wang2019learning,recht2019do}. We also evaluate on  Caltech101, Cars196, CIFAR10, CIFAR100, DTD, EuroSat, Food-101, Oxford flowers, Oxford pets, Resisc45, and Sun397 which are fine-grained classification datasets covering several different domains; see \cref{sec:datasets} for more details.

Note that our CLIP results (e.g., rows 1-3 in \cref{tab:imagenet_res,tab:finegrained_res}) differ from those presented by \citet{radford2021learning}. 
This is due to two factors.
Firstly, in several cases---e.g., for the Caltech101 dataset---\citet{radford2021learning} did not specify the dataset split they used. 
Thus we had to make guesses which do not necessarily agree with their choices; see \cref{sec:datasets} for our splits.
Secondly, we found that the implementation of \texttt{resize} in \texttt{tensorflow\_datasets}, which we used for data pre-processing, differs slightly from the \texttt{torchvision} implementation used by \citet{radford2021learning}.
This implementation difference caused large differences in performance for some datasets.

\subsection{Creating a pool of prompts}

Ideally, we would like to have a varied pool of thousands of hand-crafted prompts. 
Such a set of prompts would contain a range of generic prompts---such as `\emph{A photo of \{\}.}' and `\emph{An example of \{\}.}'---that would be useful for a many classification tasks, as well as more specific prompts---such as `\emph{A photo of \{\}, a type of flower.}' and `\emph{A cartoon of \{\}.}'---that we expect to be useful for a smaller range of tasks.
Unfortunately, no such set of prompts exists. 

In the following experiments, we simulate such a pool by combining the 27 sets of prompts designed by \citet{radford2021learning} and the prompts designed for 14 datasets by \citet{zhai2021lit}.
This leaves us with a pool of 247 unique prompts. See \cref{sec:prompt_pool} for details.
In section \ref{sec:gpt_prompts} we use the large language model, ChatGPT \citep{chatgpt}, to generate 179 additional prompt templates, resulting in a pool of 426 total templates. We then use that to study the impact of the size of the pool set on performance.

\subsection{ZPE weighted average}

\begin{table*}[t]
    \centering
    \caption{Zero-shot accuracy on ImageNet and its variants. ZPE is applied to the pool set. Note that the hand-crafted prompt sets were highly optimised and manually tuned on labeled validation sets, while our method ZPE is automatic and cheap to compute. }
    \label{tab:imagenet_res}
    \resizebox{!}{9.2em}{
    \begin{tabular}{lcccccc}
    \toprule
     & \textsc{ImageNet} &  \textsc{ImageNet-A} &  \textsc{ImageNet-R} &  \textsc{ImageNet-Sketch} &  \textsc{ImageNet-V2} &  \textbf{\textsc{Avg}} \\ \midrule
     \multicolumn{7}{c}{CLIP ViT-B/16} \\ \cmidrule{1-7}
class name                   &     63.94 &       46.01 &       74.92 &            44.12 &        57.97 &  57.39 \\
`\emph{A photo of \{\}.}'    &     66.37 &       47.47 &       73.78 &            45.84 &        60.46 &  58.78 \\
hand-crafted, equal average  &     68.31 &       49.13 &       77.31 &            47.65 &        61.83 &  60.85 \\
pool set, equal average      &     67.59 &       49.35 &       77.33 &            46.92 &        61.37 &  60.51 \\
max-logit scoring            &     67.63 &       49.37 &       77.38 &            46.95 &        61.39 &  60.55 \\
ZPE (weighted average)       &     \ul{68.56} &       \ul{49.61} &       \textbf{77.69} &            \ul{47.92} &        \textbf{62.23} &  \ul{61.20} \\
ZPE (prompt selection, ours) &     \textbf{68.60} &       \textbf{49.63} &       \ul{77.62} &            \textbf{47.99} &        \ul{62.21} &  \textbf{61.21} \\ \cmidrule{1-7}
\multicolumn{7}{c}{LiT ViT-L/16} \\ \cmidrule{1-7}
class name &78.26 &62.36 &89.80 &64.24 &71.61 &73.26 \\
`\emph{A photo of \{\}.}' &78.22 &62.43 &89.45 &63.73 &71.35 &73.03 \\
hand-crafted, equal average &78.55 &	63.09 &	90.52 &	64.90 &	72.10 &	73.83 \\
pool set, equal average &77.49 &62.07 &90.25 &63.49 &71.17 &72.89 \\
max-logit scoring &	77.86 &	62.31 &	90.47 &	63.94 &	71.31 &	73.18 \\
ZPE (weighted average) &\ul{78.90} &\ul{63.60} &\ul{90.85} &\ul{65.58} &\ul{72.43} &\ul{74.27} \\
ZPE (prompt selection, ours) &\textbf{79.26} &\textbf{63.95} &\textbf{90.91} &\textbf{65.61} &\textbf{72.59} &\textbf{74.46} \\
    \bottomrule
    \end{tabular}
    }
    \vspace{-1em}
\end{table*}

\begin{table*}[]
    \centering
    \caption{Zero-shot accuracy on fine-grained classification tasks. ZPE is applied to the pool set.}
    \label{tab:finegrained_res}
    \resizebox{!}{9.2em}{
    \begin{tabular}{lcccccccccccc}
\toprule
 & \textsc{Caltech} &  \textsc{Cars} &  \textsc{C-10} &  \textsc{C-100} &   \textsc{DTD} &  \textsc{Euro} &  \textsc{Food} &  \textsc{Flowers} &  \textsc{Pets} &  \textsc{Resisc} &  \textsc{Sun} &  \textbf{\textsc{Avg}} \\ \midrule \multicolumn{13}{c}{CLIP ViT-B/16} \\ \cmidrule{1-13}
class name                   &       77.84 &    61.60 &    87.30 &     58.59 &  44.04 &    46.90 &    86.68 &              63.57 &            81.38 &     53.74 &   60.70 &  65.67 \\
`\emph{A photo of \{\}.}'    &       82.73 &    63.45 &    88.36 &     65.49 &  42.93 &    47.85 &    88.19 &              66.84 &            87.74 &     55.96 &   59.95 &  68.13 \\
hand-crafted, equal average  &       82.82 &    \ul{64.17} &    89.10 &     65.90 &  45.64 &    51.60 &    \textbf{88.66} &              \textbf{71.23} &            \textbf{88.91} &     \textbf{65.44} &   63.87 &  70.67 \\
pool set, equal average      &       83.60 &    63.16 &    \textbf{89.56} &     65.56 &  45.96 &    \textbf{54.63} &    87.79 &              63.62 &            80.87 &     58.70 &   65.32 &  68.98 \\
max-logit scoring            &       83.56 &    63.16 &    \ul{89.55} &     65.53 &  \ul{46.28} &    \ul{54.48} &    87.81 &              63.70 &            80.87 &     59.02 &   \ul{65.39} &  69.03 \\
ZPE (weighted average)       &       \ul{84.68} &    64.13 &    89.34 &     \ul{66.40} &  \textbf{46.54} &    53.42 &    88.50 &              67.64 &            86.81 &     64.18 &   \textbf{66.15} &  \ul{70.71} \\
ZPE (prompt selection, ours) &       \textbf{85.54} &    \textbf{64.62} &    89.30 &     \textbf{66.63} &  \ul{46.28} &    53.82 &    \ul{88.61} &              \ul{70.17} &            \ul{88.72} &     \ul{64.22} &   64.70 &  \textbf{71.15} \\
 \cmidrule{1-13}
\multicolumn{13}{c}{LiT ViT-L/16}  \\ \cmidrule{1-13}
class name &83.50 &\ul{90.36} &94.86 &76.04 &55.80 &25.78 &\textbf{93.45} &78.71 &\textbf{94.74} &52.46 &69.97 &74.15 \\
`\emph{A photo of \{\}.}' &84.50 &82.07 &\ul{96.33} &77.25 &56.44 &38.97 &93.10 &\textbf{80.16} &93.38 &57.08 &70.65 &75.45 \\
hand-crafted, equal average &83.04 &86.43 &95.54 &78.32 &\textbf{60.59} &\textbf{52.19} &93.00 &79.30 &93.51 &\textbf{63.89} &69.26 &77.73 \\
pool set, equal average &83.76 &89.12 &95.64 &78.30 &57.77 &41.55 &92.65 &73.28 &90.22 &58.01 &71.13 &75.58 \\
max-logit scoring &	84.02 &	89.14 &	95.64 &	78.28 &	58.35 &	42.11 &	92.70 &	73.52 &	91.03 &	58.64 &	71.26 &	75.88 \\
ZPE (weighted average) &\ul{84.86} &90.05 &95.93 &\ul{78.98} &59.47 &48.69 &93.12 &77.75 &93.49 &62.70 &\ul{72.26} &\ul{77.94} \\
ZPE (prompt selection, ours) &\textbf{85.55} &\textbf{90.57} &\textbf{96.36} &\textbf{79.36} &\ul{60.05} &\ul{51.42} &\ul{93.32} &\ul{79.96} &\ul{93.57} &\ul{62.93} &\textbf{72.67} &\textbf{78.71} \\
\bottomrule
\end{tabular}
}
\end{table*}

\cref{tab:imagenet_res,tab:finegrained_res} show the results for using ZPE weighted averaging on our pool of prompts for ImageNet and its variants, and the fine-grained classification tasks, respectively.
On the ImageNet tasks, we see that ZPE outperforms the hand-crafted prompts across the board.
On the other hand, for the fine-grained classification tasks, performance was more mixed\footnote{
    The performance of ZPE is dependent on how well the pool of prompts matches a dataset. The datasets for which our method performs worse---e.g., Food, Flowers, and Pets---have fairly narrow domains. In \cref{sec:prompt_scores} we see that while ZPE has selected the few dataset-specific prompts available, most of the top prompts are generic.
    For datasets where we perform better, this is not the case and we tend to have a larger proportion of dataset-specific prompts with large weights.
    Thus, we attempted to use the average ZPE score as a measure of how well ZPE would perform on a given dataset. However, there was no significant relationship.
}. 
Surprisingly, for CIFAR10 and EuroSat the best performing method was an equal weighting of all of the pool prompts.
Nonetheless, ZPE beat the hand-crafted prompts on 6 of the 11 datasets, for both CLIP and LIT, performed best or second best in most cases, and performed slightly better than the hand-crafted prompts on average.
As expected ZPE performed better than naive max-logit scoring in most cases and on average.
For CLIP ViT-B/16 and LiT ViT-L/16, averaging accuracy for all 11 fine-grained datasets, ImageNet, and its four variants, ZPE gives \textbf{67.44\%} versus 66.06\% of the equal-average pool-set, and \textbf{76.79\%} versus 74.74\%, respectively. 
Comparing with a strong baseline of the hand-crafted prompts that were manually tuned over a year, which has average accuracies of 67.29\% and 76.51\% for CLIP ViT-B/16 and LiT ViT-L/16, respectively, ZPE also performs better.

Examining the top-10 prompts for ImageNet-R and Resisc45, we can see that the scores make sense given the content of the images. See \cref{sec:prompt_scores} for the top and bottom 10 prompts for all of our datasets.
\begin{tcolorbox}[colback=mplgreen!3,colframe=mplgreen!75!white,title=\textsc{ImageNet-R},left=0.5ex,right=0.5ex,top=0.5ex,bottom=0.5ex]
\fontsize{8.5}{9.}\selectfont
\emph{a drawing of a \{\}.}$\,\cdot\,$  
\emph{a drawing of the \{\}.}$\,\cdot\,$ 
\emph{itap of a \{\}.}$\,\cdot\,$ 
\emph{a sketch of a \{\}.}$\,\cdot\,$ 
\emph{a embroidered \{\}.}$\,\cdot\,$ 
\emph{a painting of a \{\}.}$\,\cdot\,$  
\emph{itap of my \{\}.}$\,\cdot\,$ 
\emph{a doodle of a \{\}.}$\,\cdot\,$ 
\emph{a painting of the \{\}.'}
\emph{itap of the \{\}.}$\,\cdot\,$  
\end{tcolorbox}
\begin{tcolorbox}[colback=mplgreen!3,colframe=mplgreen!75!white,title=\textsc{Resisc45},left=0.5ex,right=0.5ex,top=0.5ex,bottom=0.5ex]
\fontsize{9.}{9.}\selectfont
\emph{satellite view of a \{\}.}$\,\cdot\,$
\emph{satellite view of the \{\}.}$\,\cdot\,$ 
\emph{satellite view of \{\}.}$\,\cdot\,$
\emph{satellite photo of a \{\}.}$\,\cdot\,$
\emph{satellite photo of the \{\}.}$\,\cdot\,$
\emph{a centered satellite photo of a \{\}.}$\,\cdot\,$
\emph{a satellite image of \{\}.}$\,\cdot\,$
\emph{satellite photo of \{\}.}$\,\cdot\,$
\emph{a centered satellite photo of the \{\}.}$\,\cdot\,$ 
\emph{a satellite photo of \{\}.}
\end{tcolorbox}

\subsection{ZPE prompt selection}

In addition to using the softmax function to down-weight the bad prompts using \cref{eq:softmax_prompt_ensemble}, we can select a set of top prompts and use \cref{eq:masked_prompt_ensemble} for prompt ensembling.
For prompt selection, we need to choose a proper hyper-parameter $\tau$.
For ImageNet and its variants, since the dataset contains a large set of diverse classes, a diverse set of prompts are needed for good performance, while for the fine-grained datasets, less diverse but more domain specific prompts fit the task better. Therefore, we use $\tau=0.5$ for ImageNet and its variants, and $\tau=2.0$ for all fine-grained datasets. These values were chosen by sweeping over $\tau \in \{0.1, 0.2, 0.3, 0.4, 0.5, 1.0, 1.5, 1.8, 2.0, 2.5\}$ and choosing the best values according to the average classification performance across datasets. This is similar to how hyper-parameters are chosen in previous few-shot settings \citep{dosovitskiy21image,allingham2022sparse}.

\cref{tab:imagenet_res,tab:finegrained_res} %
provide comparisons between our prompt selection and weighted average methods, for the ImageNet (with variants) and fine-grained classification tasks, respectively.
We see that for ImageNet and variants, both methods perform very similarly, with the prompt selection providing slightly better results on average. 
On the other hand, for the fine-grained tasks, prompt selection is better on 8 and all of the 11 datasets, for CLIP ViT-B/16 and LiT ViT-L/16 respectively. We also see that the average accuracy is higher by a more significant margin for both models.
This result agrees with our intuition that fine-grained tasks will need fewer prompts than general tasks like ImageNet, and will thus benefit more from prompt selection.
For CLIP ViT-B/16 and LiT ViT-L/16, averaging accuracy for all of our datasets, comparing ZPE prompt selection and the equal-average pool-set gives \textbf{67.73\%} versus 66.06\% and \textbf{77.38\%} versus 74.74\%, respectively.
Comparing with the strong baseline of hand-crafted prompts---with 67.29\%, and 76.51\% on average for CLIP ViT-B/16 and LiT ViT-L/16, respectively---ZPE performs better again.

\subsection{Ablation studies and sensitivity analyses}

In this section we perform ablations studies to show that each component of our algorithm is responsible for good performance. 
We also investigate the generalization of our algorithm by performing sensitivity analyses.
We also perform our investigation using ZPE weighted average, in addition to prompt selection, to avoid a potential confounding factor in the selection of $\tau$.

\subsubsection{Normalization schemes ablation}

\cref{tab:norm} compares zero-shot performance for various normalization schemes. 
We see that our combination of $E_{\text{pretrain}}$ and $E_{\text{test}}$ normalization works best in most cases, providing a 0.54\% and 0.61\% average increase in zero-shot accuracy compared to no normalization, for the weighted average and prompt selection, respectively.
We also see that while $E_{\text{pretrain}}$ normalization does not seem to hurt aggregate performance in most cases, $E_{\text{test}}$ normalization performed in isolation can hurt, despite helping when combined with $E_{\text{pretrain}}$.
$E_{\text{pretrain}}$ seems to be the most important component.
Finally, we also investigated a variant of $E_{\text{pretrain}}$ normalization in which we removed the impact of the class names by taking the expectation over both the images and classes.
However, this scheme tends to perform worse than image-only normalization.

\begin{table}[htbp]
   \vspace{-1em} 
   \centering
    \caption{Ablation study for normalization schemes. Zero-shot accuracy for CLIP ViT-B/16 on ImageNet, as well as on average for the ImageNet variants, the fine-grained classification datasets, and all of our datasets. 
    $E_{\text{pretrain}}^*$ is a variant in which we take the expectation over both the images and classes.}
    \label{tab:norm}
    \resizebox{!}{8.2em}{
    \begin{tabular}{lcccc}
    \toprule
     &  \textsc{INet} &  \textsc{Variants} &  \textsc{Fine} &  \textbf{\textsc{All}} \\ \midrule
     \multicolumn{5}{c}{weighted average} \\ \cmidrule{1-5}
none & 68.17 &  59.30 &  69.99 &  66.92 \\
$E_{\text{pretrain}}$ & \textbf{68.64} &     \ul{59.31} &  \ul{70.70} &  \ul{67.42} \\
$E_{\text{pretrain}}^*$ & \ul{68.62} &     \ul{59.31} &  70.33 &  67.17 \\
$E_{\text{test}}$ & 68.45 &     59.23 &  70.11 &  67.00 \\
both (ZPE) & 68.56 &     \textbf{59.36} &  \textbf{70.71} &  \textbf{67.44} \\ \cmidrule{1-5}
\multicolumn{5}{c}{prompt selection} \\ \cmidrule{1-5}
none & 68.24 & \textbf{59.37} &  70.30 &  67.15 \\
$E_{\text{pretrain}}$ & \ul{68.64} &     59.25 &  \ul{71.13} &  \ul{67.69} \\
$E_{\text{pretrain}}^*$ & \textbf{68.66} &     59.26 &  70.67 &  67.39 \\
$E_{\text{test}}$ & 68.54 &     59.10 &  70.29 &  67.09 \\
both (ZPE) & 68.60 &     \ul{59.36} &  \textbf{71.15} &  \textbf{67.73} \\
    \bottomrule
    \end{tabular}
    }
\end{table}

\subsubsection{Weighting schemes ablation}
\label{sec:weight_with_softmax}

\cref{tab:weighting} compares zero-shot performance for three weighting schemes. 
We see that our method of taking the softmax of the scores provides the best performance on average, and always performs better than using the raw scores, particularly in the weighted average case, as expected.
As a sanity check, we also compare with raising the scores to the power of 10. 
As expected, this also tends to improve performance relative to the raw scores, confirming the benefit of a relative up-weighing of the good prompts.

\begin{table}[htbp]
    \centering
    \caption{Ablation study for weighting schemes. Zero-shot accuracy for CLIP ViT-B/16 on ImageNet, as well as on average for the ImageNet variants, the fine-grained classification datasets, and all of our datasets.}
    \label{tab:weighting}
    \resizebox{!}{6.2em}{
    \begin{tabular}{lcccc}
    \toprule
     &  \textsc{INet} &  \textsc{Variants} &  \textsc{Fine} &  \textbf{\textsc{All}} \\ \midrule
     \multicolumn{5}{c}{weighted average} \\ \cmidrule{1-5}
$\mathtt{scores}$ & 67.74 &     58.79 &  69.13 &  66.18 \\
$\mathtt{scores}^{10}$ & \ul{68.35} &     \ul{59.32} &  \ul{70.55} &  \ul{67.30} \\
$\mathrm{softmax}$ (ZPE) & \textbf{68.56} &     \textbf{59.36} &  \textbf{70.71} &  \textbf{67.44} \\ \cmidrule{1-5}
\multicolumn{5}{c}{prompt selection} \\ \cmidrule{1-5}
$\mathtt{scores}$ & 68.55 &     59.31 &  71.12 &  67.70 \\
$\mathtt{scores}^{10}$ & \textbf{68.61} &     \textbf{59.37} &  \ul{71.13} &  \ul{67.72} \\
$\mathrm{softmax}$ (ZPE) & \ul{68.60} &     \ul{59.36} &  \textbf{71.15} &  \textbf{67.73} \\
    \bottomrule
    \end{tabular}
    }
\end{table}

\subsubsection{Model architecture sensitivity}

We investigate the sensitivity of our method to the architecture of the underlying text-image model. 
\cref{tab:models} shows performance of ZPE relative to hand-crafted prompts, for a range of CLIP and LiT model architectures. 
We see that ZPE, especially with prompt selection, improves on the equal-average pool-set baseline and the hand-crafted prompts, performing better on average in most cases. 
Since the hand-crafted prompts were designed for CLIP rather than LiT we expect that ZPE will provide larger performance gains for the LiT models, which is indeed the case.
This showcases a key benefit of our method, namely that we avoid the need to hand-tune the set of prompts for each task and model, and is in-line with the intuition of \citet{pham2021combined} who hypothesised that the prompts of \citet{radford2021learning} might not be optimal for other text-image models.

\begin{table}[h!]
    \centering
    \caption{Ablation study for different kinds of text-image models. 
    Zero-shot accuracy on ImageNet, as well as on average for the ImageNet variants, the fine-grained classification datasets, and all of our datasets. 
    }
    \label{tab:models}
    \resizebox{\linewidth}{!}{
\begin{tabular}{lcccc}
\toprule
     & \multicolumn{1}{l}{\textsc{INet}}               & \multicolumn{1}{l}{\textsc{Variants}}           & \multicolumn{1}{l}{\textsc{Fine}}              & \multicolumn{1}{l}{\textbf{\textsc{All}}}                \\ \toprule
\multicolumn{5}{c}{CLIP ResNet-50}                                                                                                                                                                \\ \midrule
hand-crafted, equal average  & 59.48          & 42.52          & \underline{59.36}    & \underline{55.15}    \\
pool set, equal average      & 58.24          & 42.17          & 56.04          & 52.71          \\
ZPE (weighted average) & \underline{59.68}    & \textbf{42.97} & 58.79          & 54.89          \\
ZPE (prompt selection, ours) & \textbf{59.90} & \underline{42.87}    & \textbf{59.64} & \textbf{55.46} \\ \cmidrule{1-5}
\multicolumn{5}{c}{CLIP ResNet-101}                                                                                                                                                              \\ \cmidrule{1-5}
hand-crafted, equal average  & 62.47          & 48.57          & \underline{62.33}    & \underline{58.90}    \\
pool set, equal average      & 61.56          & 48.16          & 59.86          & 57.04          \\
ZPE (weighted average) & \underline{62.66}    & \underline{48.81}    & 61.92          & 58.69          \\
ZPE (prompt selection, ours) & \textbf{62.80} & \textbf{48.86} & \textbf{62.66} & \textbf{59.21} \\ \cmidrule{1-5}
\multicolumn{5}{c}{CLIP ViT-B/32}                                                                                                                                                                \\ \cmidrule{1-5}
hand-crafted, equal average  & 62.95          & 49.44          & 67.59          & 62.76          \\
pool set, equal average      & 61.73          & 48.97          & 65.34          & 61.02          \\
ZPE (weighted average) & \underline{63.16}    & \underline{49.66}    & \underline{67.69}    & \underline{62.90}    \\
ZPE (prompt selection, ours) & \textbf{63.31} & \textbf{49.76} & \textbf{68.05} & \textbf{63.18} \\ \cmidrule{1-5}
\multicolumn{5}{c}{CLIP ViT-B/16}                                                                                                                                                                \\ \cmidrule{1-5}
hand-crafted, equal average  &     68.31 &     58.98 &  70.67 &  67.29 \\
pool set, equal average      &     67.59 &     58.74 &  68.98 &  66.06 \\
ZPE (weighted average)       &     \ul{68.56} &     \textbf{\ul{59.36}} &  \ul{70.71} &  \ul{67.44} \\
ZPE (prompt selection, ours) &     \textbf{68.60} &     \textbf{\ul{59.36}} &  \textbf{71.15} &  \textbf{67.73} \\ \cmidrule{1-5}
\multicolumn{5}{c}{CLIP ViT-L/14}                                                                                                                                                                \\ \cmidrule{1-5}
hand-crafted, equal average  & 75.36          & 71.72          & \underline{77.40}    & \underline{75.85}    \\
pool set, equal average      & 74.77          & 71.41          & 74.60          & 73.82          \\
ZPE (weighted average) & \underline{75.58}    & \underline{72.01}    & 77.18          & 75.79          \\
ZPE (prompt selection, ours) & \textbf{75.62} & \textbf{72.02} & \textbf{77.67} & \textbf{76.13} \\ \cmidrule{1-5}
\multicolumn{5}{c}{LiT ViT-B/32}                                                                                                                                                                 \\ \cmidrule{1-5}
hand-crafted, equal average  & 68.13          & 55.25          & 70.19          & 66.33          \\
pool set, equal average      & 66.93          & 54.51          & 68.55          & 64.94          \\
ZPE (weighted average) & \underline{68.60}    & \underline{55.67}    & \underline{70.81}    & \underline{66.89}    \\
ZPE (prompt selection, ours) & \textbf{68.88} & \textbf{55.72} & \textbf{71.78} & \textbf{67.58} \\ \cmidrule{1-5}
\multicolumn{5}{c}{LiT ViT-B/16}                                                                                                                                           \\ \cmidrule{1-5}
hand-crafted, equal average  & 73.24          & 64.61          & 73.03          & 70.94          \\
pool set, equal average      & 72.29          & 63.70          & 70.47          & 68.89          \\
ZPE (weighted average) & \underline{73.93}    & \underline{64.95}    & \underline{73.17}    & \underline{71.16}    \\
ZPE (prompt selection, ours) & \textbf{74.02} & \textbf{65.14} & \textbf{73.88} & \textbf{71.71} \\ \cmidrule{1-5}
\multicolumn{5}{c}{LiT ViT-L/16}                                                                                                                                                                 \\ \cmidrule{1-5}
hand-crafted, equal average  &  78.55 &	72.65 &	77.73 &	76.51          \\
pool set, equal average      & 77.49 &	71.74 &	75.58 &	74.74          \\
ZPE (weighted average) & \underline{78.90}    & \underline{73.11}    & \underline{77.94}    & \underline{76.79}    \\
ZPE (prompt selection, ours) & \textbf{79.26} & \textbf{73.27} & \textbf{78.71} & \textbf{77.38} \\\bottomrule
\end{tabular}

}
\vspace{-1em}
\end{table}

\subsubsection{Other ablations and sensitivity analyses} 

In \cref{sec:gpt_prompts}, we investigate the impact of the size of the pool set on performance. We used ChatGPT \cite{chatgpt} to generate additional 179 prompts which results in a total of 427 prompts. We see that ZPE still outperforms the hand-crafted equal-average method. 
We also study the effect of the number of random images used for estimating $\mathbb{E}_{\mathtt{image}\sim \mathcal{D}_{\text{pretrain}}}$ and the number of test images used for estimating $\mathbb{E}_{\mathtt{image}\sim \mathcal{D}_{\text{test}}}$. 
We see that ZPE is very robust to those sample sizes. ZPE scores can be reliably estimated using as few as 5k random images and as little as 10\% of test data. See \cref{sec:random_n_sample,sec:test_n_sample} for details.

\section{Related Work} \label{sec:related_work}

\paragraph{Prompt engineering in large language models.} Prompting of large language models has been shown to improve their performance on downstream tasks via `in-context learning' \cite{brown2020language}. Prompt engineering---i.e., the process of finding a good set of training examples to demonstrate the task and prepend them to the final test task---has since become an integral component of machine learning pipelines involving large language models.
The automatic generation of prompts has also been explored in a range of works including \citet{shin2020autoprompt,gao2021making,zhou2022large}.
Of particular relevance to our work is \citet{rubin2022learning} who treat prompt engineering as a retrieval problem given a set of candidate prompts. 
Similarly, \citet{sorenson2022information} is relevant to our work for their use of an unsupervised learning objective.
The above methods for language models can not be directly applied to text-image models, since the `prompts' in the text-image case refer to the templates to compose with the class name, rather than the context examples to prepend to the test task.

\paragraph{Prompt engineering in text-image models.} Much like large language models, text-image models benefit from prompt engineering. In particular, \citet{radford2021learning} showed that prompt engineering provides significant accuracy gains for zero-shot classifiers. However, \citet{radford2021learning} hand-crafted prompts for each of their downstream classification tasks.  
In contrast, CoOp \citep{zhou2022learning} and CoCoOp \citep{zhou2022conditional} automatically learn a continuous context vector for each downstream task which acts as a prompt. Unlike the prompts of \citet{radford2021learning}, the learned context vectors are not always interpretable.
Furthermore, CoOp and CoCoOp require a small amount of labeled training data for each class to learn the context, which conflicts the zero-shot applicability of the classifiers.
Additionally, the context vectors are poorly generalizable from one dataset to another, suffering from overfitting. 
The closest work to ours is test-time prompt tuning (TPT) \citep{shu2022test}. 
Similar to CoOp, TPT learns context vectors. However, TPT uses an unsupervised entropy minimization objective based on the idea that a good prompt vector should yield consistent classification of an image under different augmentations.
It is worth noting that TPT is much more complex than ZPE requiring optimization at test-time and having many more hyper-parameters to tune, including the choice of data augmentation strategy, a filtering threshold, and optimizer related hyper-parameters.
ZPE prompt selection instead has only one hyper-parameter $\tau$, and doesn't require any optimization or learning, which makes it simpler and cheaper to deploy.
ZPE also provides an interpretable output while TPT does not.

\paragraph{Reducing bias via normalization.}

Using expected value to normalize the raw score for the purpose of reducing bias is a commonly used technique. 
Term Frequency–Inverse Document Frequency (TF-IDF), which is widely used for query rankings in information retrieval \cite{jones1972statistical}. TF-IDF takes the difference between the raw word frequency in a document and the expected word frequency estimated via documents in a corpus. Similar to our approach, the normalization is to down-weight the frequent but non-meaning words like `the'.
\citet{ren2019likelihood, ren2021simple, ren2022out} discover that the raw likelihood score from deep generative models can be biased towards background statistics, and proposed to use a expected likelihood to correct for the bias which results in significant improvement on out-of-distribution detection performance.
In addition, in medical studies, the ratio of observed to expected mortality rate (O/E) is also often used as an indicator of quality of treatment \cite{best1994ratio,galvan2015observed}.

\section{Discussion and Conclusion}

We have introduced zero-shot prompt ensembling, a technique for improving the zero-shot accuracy of text-image models without the need for manual prompt engineering. 
We identified and addressed several pathologies involved in a naive implementation of our algorithm. 
Our algorithm outperforms the equal-average pool-set baseline and even the strong baseline of hand-crafted prompts
, while remaining simple to implement---with no training required---and essentially free to apply.
Nonetheless, there is room for future work. 
In particular, the following limitations could be addressed:
\begin{itemize}%
    \item ZPE assumes access to a large and varied pool of high-quality prompt templates, which to the best of our knowledge has not been collected.  Constructing such a pool of prompts could improve the performance of our algorithm.
    \item ZPE scores prompts independently, without considering combinations of prompts. Scoring combinations of prompts could yield further gains.
    \item Scoring is done per dataset rather than per image. However, a prompt that is good for one image might not be useful for another. For example, `\emph{A photo of a \{\}, a type of cat.}' would likely not be useful for an image of a dog. Furthermore, \citet{zhou2022learning} showed that learning context vectors on a per-image, rather than per-dataset, basis can provide improved performance. Our initial investigation, see \cref{sec:per_example_scores}, shows that per-example scoring can indeed lead to improved performance, suggesting that this might be a fruitful direction for future work.
\end{itemize}
Finally, using a small amount of labeled data---i.e., in a few-shot setting---to select prompts could be investigated.

\section*{Acknowledgements}

The authors would like to thank Clara Huiyi Hu and Sharat Chikkerur for their feedback on this draft. The authors would also like to thank Ed Chi, Rodolphe Jenatton, Efi Kokiopoulou, Mark Collier, Basil Mustafa, and Jannik Kossen for their feedback throughout the project.

\clearpage

\bibliography{references}
\bibliographystyle{icml2023}

\appendix
\onecolumn

\section{Additional Experimental Results} \label{sec:more_results}

\subsection{Sensitivity to the size of the pool set} \label{sec:gpt_prompts}

We investigate the impact of the size of the pool set on performance. 
To do this, we use two additional pools of prompts. The first is the set of 80 prompts designed by \citet{radford2021learning} for ImageNet. We constructed the second set by using ChatGPT \citep{chatgpt} to create additional prompts by filling in the following templates:
\begin{itemize}
    \item `\emph{A photo of a \{\}, a type of \texttt{XXX}.}', where \texttt{XXX} was replaced with categories of objects that could be found in a photo, for example `\emph{insect}', `\emph{fish}', and `\emph{tree}',
    \item `\emph{A \texttt{YYY} photo of a \{\}.}', where \texttt{YYY} was replaced with adjectives that could describe an image, for example `\emph{panoramic}', `\emph{close-up}', and `\emph{wide-angle}',
    \item `\emph{A \texttt{ZZZ} of a \{\}.}', where \texttt{ZZZ} was replaced with different mediums, for example `\emph{print}', `\emph{engraving}', and  `\emph{etching}'.
\end{itemize}
We created an additional 179 prompt templates which resulted in a pool of 426 total templates. See \cref{sec:prompt_pool} for further details.

\cref{tab:pool_size} shows the performance of ZPE with the different prompt pools. In both the weighted average and prompt selection cases, we see that the additional prompts reduce the average performance below 247-prompt pool but not below the 80-prompt pool. As expected, the 80-prompt pool, performs well on ImageNet but less well on the fine-grained datasets. These results indicate that both quantity and quality of the prompt pool are important. The 80-prompt poll seems to not have enough diverse prompts, while the 179 ChatGPT generated prompts in 426-prompt pool seem to bring the quality of the pool down.
In most cases, ZPE performs competitively or better than the equal-average hand-crafted prompts. 
We also investigate whether ZPE scores can improve the performance of the hand designed prompts provided by \citet{radford2021learning}.
We see that ZPE weighting outperforms the naive weighting of the hand designed prompts. 
Considering that the hand-crafted prompts are highly optimised---having being tuned over the course of a year\footnote{More details in \scriptsize
\url{https://github.com/openai/CLIP/blob/main/notebooks/Prompt_Engineering_for_ImageNet.ipynb}.}---and that ZPE is automatic and cheap to compute, providing a performance boost almost for free, the improvements are substantial.

\begin{table}[htbp]
    \centering
    \caption{ZPE sensitivity to pool size. Zero-shot accuracy for CLIP ViT-B/16 on ImageNet, as well as on average for the ImageNet variants, the fine-grained classification datasets, and all of our datasets. \textcolor{mplgray}{Gray} indicates that ZPE performs worse than the hand-crafted prompts.}
    \label{tab:pool_size}
    \begin{tabular}{lcccc}
    \toprule
     &  \textsc{INet} &  \textsc{Variants} &  \textsc{Fine} &  \textbf{\textsc{All}} \\ \midrule
     hand-crafted, equal average & 68.31 &     58.98 &  70.67 &  67.29 \\ 
     hand-crafted, ZPE weights & \textbf{68.57} &	\textbf{59.26} &	\textbf{70.74} &	\textbf{67.43} \\\cmidrule{1-5}
          \multicolumn{5}{c}{weighted average} \\ \cmidrule{1-5}
ZPE (80 prompts) &\textbf{68.57} & \ul{59.26} & \textcolor{mplgray}{70.29} & \textcolor{mplgray}{67.13} \\
ZPE (247 prompts) & \ul{68.56} &\textbf{59.36} & \textbf{70.71} & \textbf{67.44} \\
ZPE (426 prompts) &68.40 &59.24 & \textcolor{mplgray}{\ul{70.54}} & \ul{67.30} \\ \cmidrule{1-5}
     \multicolumn{5}{c}{prompt selection} \\ \cmidrule{1-5}
ZPE (80 prompts) & \ul{68.38} &{59.08} &{\textcolor{mplgray}{70.58}} &\textcolor{mplgray}{67.26} \\
ZPE (247 prompts) &\textbf{68.60} &\textbf{59.36} &\textbf{71.15} &\textbf{67.73} \\
ZPE (426 prompts) & 68.37 & \ul{59.21} &\ul{70.96} &\ul{67.54} \\
    \bottomrule
    \end{tabular}
\end{table}

\subsection{Sensitivity to the number of random images} \label{sec:random_n_sample}

In order to de-bias our scores, we use a set of random images that cover a wide range of natural images such that we have an accurate approximation for $\mathbb{E}_{\mathtt{image}\sim \mathcal{D}_{\text{pretrain}}}\left[\mathbf{z}_\text{img} \cdot \mathbf{z}_\text{txt}\right]$. 
Here we investigate the sensitivity of ZPE to the number of images used to approximate this expectation. 

\cref{tab:background_size} compares the performance with first 5k, 10k, and 20k images from the LAION400m dataset \citep{schuhmann2021laion}. We see that ZPE is very robust to the number of random images.

\begin{table}[htbp]
\vspace{-1em}
    \centering
    \caption{ZPE sensitivity to the number of random images used for estimating $\mathbb{E}_{\mathtt{image}\sim \mathcal{D}_{\text{pretrain}}}$. Zero-shot accuracy for CLIP ViT-B/16 on ImageNet, as well as on average for the ImageNet variants, the fine-grained classification datasets, and all of our datasets.}
    \label{tab:background_size}
    \begin{tabular}{lcccc}
    \toprule
     &  \textsc{INet} &  \textsc{Variants} &  \textsc{Fine} &  \textbf{\textsc{All}} \\ \midrule
     \multicolumn{5}{c}{weighted average} \\ \cmidrule{1-5}
5k & 68.55 &     59.36 &  70.71 &  67.44 \\
10k & 68.56 &     59.36 &  70.71 &  67.44 \\
20k & 68.56 &     59.36 &  70.71 &  67.44 \\ \cmidrule{1-5}
\multicolumn{5}{c}{prompt selection} \\ \cmidrule{1-5}
5k & 68.60 &     59.37 &  71.15 &  67.74 \\
10k & 68.60 &     59.37 &  71.15 &  67.73 \\
20k & 68.60 &     59.36 &  71.15 &  67.73  \\
    \bottomrule
    \end{tabular}
    \vspace{-1em}
\end{table}

\subsection{Sensitivity of the number of test images for prompt score estimation}
\label{sec:test_n_sample}

We used all the images in the test dataset to estimate $\mathbb{E}_{\mathtt{image}\sim \mathcal{D}_{\text{test}}}\left[\mathbf{z}_\text{img} \cdot \mathbf{z}_\text{txt}\right]$. 
In this section, we study how sensitive is the estimation to the test sample size. 
Instead of using all images in the test set, we use 10\%, 20\%, 50\% of the test samples to estimate $\mathbb{E}_{\mathtt{image}\sim \mathcal{D}_{\text{test}}}$. 
\cref{tab:test_size} shows the zero-shot classification accuracy when using partial test dataset. We see that ZPE is very robust to the number of test images used for score estimation.

\begin{table}[htbp]
\vspace{-1em}
    \centering
    \caption{ZPE sensitivity to the percentage of test images used for estimating $\mathbb{E}_{\mathtt{image}\sim \mathcal{D}_{\text{test}}}$. Zero-shot accuracy for CLIP ViT-B/16 on ImageNet, as well as on average for the ImageNet variants, the fine-grained classification datasets, and all of our datasets.}
    \label{tab:test_size}
    \begin{tabular}{lcccc}
    \toprule
     &  \textsc{INet} &  \textsc{Variants} &  \textsc{Fine} &  \textbf{\textsc{All}} \\ \midrule
     \multicolumn{5}{c}{weighted average} \\ \cmidrule{1-5}
10\% &     68.55 &     59.35 &  70.71 &  67.44 \\
20\% &     68.56 &     59.36 &  70.71 &  67.44 \\
50\% &     68.56 &     59.36 &  70.71 &  67.44 \\
100\% &     68.56 &     59.36 &  70.71 &  67.44 \\ \cmidrule{1-5}
\multicolumn{5}{c}{prompt selection} \\ \cmidrule{1-5}
10\% &     68.59 &     59.39 &  71.16 &  67.75 \\
20\% &     68.60 &     59.37 &  71.12 &  67.72 \\
50\% &     68.60 &     59.37 &  71.15 &  67.74 \\
100\% &     68.60 &     59.36 &  71.15 &  67.73 \\
    \bottomrule
    \end{tabular}
    \vspace{-1em}
\end{table}

\subsection{Initial investigation into per-example scoring} \label{sec:per_example_scores}

\cref{tab:per_example_scores} compares the performance of per-example scoring with per-dataset scoring (i.e., as used for the results in the main text).
In the case of the hand-crafted prompts, we see that per-dataset scoring performs better across the board. 
However, the per-example scoring does perform better than the equal average. 

On the other hand, for the pool set, we see that the per-example scoring can perform better than per-dataset in some cases, and indeed, on average. 
In particular, per-example seems to work better for the fine-grained classification datasets. 

We also see that, while softmax scoring and normalisation are still important for per-example scores, their removal is less impactful than in the per-dataset case. 

\begin{table}[]
    \centering
    \caption{Results for per-example scoring. Zero-shot accuracy for CLIP ViT-B/16 on ImageNet, as well as on average for the ImageNet variants, the fine-grained classification datasets, and all of our datasets.}
    \label{tab:per_example_scores}
    \begin{tabular}{lcccc}
    \toprule
     &  \textsc{INet} &  \textsc{Variants} &  \textsc{Fine} &  \textbf{\textsc{All}} \\
    \midrule
    hand crafted, equal average            &     \ul{68.31} &     58.98 &  \ul{70.67} &  67.29 \\
    hand crafted, ZPE weights, per-dataset &     \textbf{68.57} &     \textbf{59.26} &  \textbf{70.74} &  \textbf{67.43} \\
    hand crafted, ZPE weights, per-example &     68.11 &     \ul{59.10} &  \ul{70.67} &  \ul{67.31} \\ \cmidrule{1-5}
    pool set, equal average            &     67.59 &     58.74 &  68.98 &  66.06 \\
pool set, ZPE weights, per-dataset &     \textbf{68.56} &     \textbf{59.36} &  \ul{70.71} &  \ul{67.44} \\
pool set, ZPE weights, per-example &     \ul{67.97} &     \ul{59.34} &  \textbf{71.01} &  \textbf{67.60} \\ \cmidrule{1-5}
pool set, ZPE weights, per-dataset, no softmax &     \ul{67.74} &     \ul{58.79} &  \ul{69.13} &  \ul{66.18} \\
pool set, ZPE weights, per-example, no softmax &     \textbf{67.84} &     \textbf{58.88} &  \textbf{69.37} &  \textbf{66.37} \\ \cmidrule{1-5}
pool set, ZPE weights, per-dataset, no norm &     \textbf{68.17} &      \ul{59.30} &  \ul{69.99} &  \ul{66.92} \\
pool set, ZPE weights, per-example, no norm &     \ul{67.92} &      \textbf{59.40} &  \textbf{70.71} &  \textbf{67.42} \\
    \bottomrule
    \end{tabular}
\end{table}

\clearpage
\section{Additional Experimental Details} \label{sec:experimental_details}

\subsection{Dataset Details} \label{sec:datasets}

We used the 16 datasets from \citet{radford2021learning} with entries in \texttt{tensorflow\_datasets}. We used the test, validation, and train splits as available, in that order of preference. \cref{tab:datasets} provides the details for each dataset.

\begin{table}[htbp]
    \centering
    \caption{Details for the datasets used in our experiments.}
    \label{tab:datasets}
    \begin{tabular}{cccc}
    \toprule
        \textsc{Dataset} & \textsc{Classes} & \textsc{Split} & \textsc{Reference}\\ \midrule
    ImageNet & 1000 & validation & \citep{russakovsky2015imagenet}  \\ \cmidrule{1-4}
    ImageNet-R & 200 & test & \citep{hendrycks2021many}  \\
    ImageNet-A & 200 & test & \citep{hendrycks2021natural} \\
    ImageNet-Sketch & 1000 & test & \citep{wang2019learning} \\
    ImageNet-V2 & 1000 & test & \citep{recht2019do} \\ \cmidrule{1-4}
    Caltech101 & 102 & test & \citep{FeiFei2004LearningGV} \\
    Cars196 & 196 & test & \citep{krause20133d} \\
    CIFAR10 & 10 & test & \citep{Krizhevsky09learningmultiple} \\
    CIFAR100 & 100 & test & \citep{Krizhevsky09learningmultiple} \\
    DTD & 47 & test & \citep{cimpoi2014describing} \\
    EuroSat & 10 & train & \citep{helber2017eurosat} \\
    Food101 & 101 & validation & \citep{bossard14} \\
    Oxford Flowers & 102 & test & \citep{Nilsback08}\\
    Oxford Pets & 37 & test & \citep{parkhi12a} \\
    Resisc45 & 45 & train & \citep{Cheng_2017} \\
    Sun397 & 397 & test &  \citep{Xiao:2010} \\
    \bottomrule
    \end{tabular}
\end{table}

\subsection{Implementation Details} \label{sec:implementation}

Our code makes use of \texttt{uncertainty\_baselines} \citep{nado2021uncertainty} and can be found at \url{https://github.com/google/uncertainty-baselines/tree/main/experimental/multimodal}.
We provide a notebook for reproducing all CLIP results from this paper.
Our models are implemented in \texttt{jax} \citep{jax2018github} with \texttt{flax} \citep{flax2020github}. Our LiT implementation is from \texttt{big\_vision} \citep{big_vision}.
We use the public CLIP weights provided by \citet{radford2021learning} at \url{https://github.com/openai/CLIP/blob/main/clip/clip.py}. 
We use private LiT weights provided by \citet{zhai2021lit}.

\twocolumn

\section{Per-dataset Prompt Scores} \label{sec:prompt_scores}
The following table contains the top and bottom 10 prompts in the pool set, and the corresponding ZPE scores, for each of our datasets, for CLIP ViT-B/16.

\fontsize{8.}{9.5}\selectfont
\tablefirsthead{\toprule
\textsc{No.} & \textsc{Prompt} & \ \textsc{Score} \\ \midrule}
\tablehead{\toprule
\textsc{No.} & \textsc{Prompt} & \ \textsc{Score} \\ \midrule}

\begin{supertabular}{rlc}
\multicolumn{3}{c}{\textsc{ImageNet}}  \\ \cmidrule{1-3}
1 & `\emph{itap of a \{\}.}' & 0.0585 \\
2 & `\emph{itap of the \{\}.}' & 0.0387 \\
3 & `\emph{itap of my \{\}.}' & 0.0373 \\
4 & `\emph{a black and white photo of a \{\}.}' & 0.0234 \\
5 & `\emph{a high contrast photo of a \{\}.}' & 0.0199 \\
6 & `\emph{a photo of a large \{\}.}' & 0.0183 \\
7 & `\emph{a photo of the large \{\}.}' & 0.0179 \\
8 & `\emph{a black and white photo of the \{\}.}' & 0.0172 \\
9 & `\emph{a low contrast photo of a \{\}.}' & 0.0169 \\
10 & `\emph{a example of a person \{\}.}' & 0.0145 \\
\multicolumn{3}{c}{\vdots} \\
238 & `\emph{the closest shape in this rendered image is \{\}.}' & 0.0001 \\
239 & `\emph{something rotated at \{\}}' & 0.0001 \\
240 & `\emph{they look \{\}.}' & 0.0001 \\
241 & `\emph{the nearest shape in this image is \{\}.}' & 0.0001 \\
242 & `\emph{there are \{\} shapes in the image.}' & 0.0001 \\
243 & `\emph{a video of the person during \{\}.}' & 0.0001 \\
244 & `\emph{a fundus image with signs of \{\}}' & 0.0001 \\
245 & `\emph{a photo of the person during \{\}.}' & 0.0001 \\
246 & `\emph{a zoomed in photo of a "\{\}" traffic sign.}' & 0.0001 \\
247 & `\emph{there are \{\} objects in the image.}' & 0.0001 \\
\cmidrule{1-3}
\multicolumn{3}{c}{\textsc{ImageNet-A}}  \\ 
\cmidrule{1-3}
1 & `\emph{itap of a \{\}.}' & 0.0858 \\
2 & `\emph{itap of the \{\}.}' & 0.0588 \\
3 & `\emph{itap of my \{\}.}' & 0.0504 \\
4 & `\emph{a high contrast photo of a \{\}.}' & 0.0168 \\
5 & `\emph{a low contrast photo of a \{\}.}' & 0.0164 \\
6 & `\emph{a photo of a large \{\}.}' & 0.0151 \\
7 & `\emph{a photo of the large \{\}.}' & 0.0128 \\
8 & `\emph{a black and white photo of a \{\}.}' & 0.0128 \\
9 & `\emph{a dark photo of a \{\}.}' & 0.0112 \\
10 & `\emph{a cropped photo of a \{\}.}' & 0.0111 \\
\multicolumn{3}{c}{\vdots} \\
238 & `\emph{a zoomed in photo of a "\{\}" traffic sign.}' & 0.0004 \\
239 & `\emph{a video of the person during \{\}.}' & 0.0003 \\
240 & `\emph{a rendered image of \{\} objects.}' & 0.0003 \\
241 & `\emph{the closest shape in this rendered image is \{\}.}' & 0.0003 \\
242 & `\emph{a face that looks \{\}.}' & 0.0003 \\
243 & `\emph{they look \{\}.}' & 0.0003 \\
244 & `\emph{there are \{\} shapes in the image.}' & 0.0003 \\
245 & `\emph{a fundus image with signs of \{\}}' & 0.0003 \\
246 & `\emph{a photo of the person during \{\}.}' & 0.0002 \\
247 & `\emph{there are \{\} objects in the image.}' & 0.0002 \\
\cmidrule{1-3}
\multicolumn{3}{c}{\textsc{ImageNet-R}}  \\ \cmidrule{1-3}
1 & `\emph{a drawing of a \{\}.}' & 0.0410 \\
2 & `\emph{a drawing of the \{\}.}' & 0.0315 \\
3 & `\emph{itap of a \{\}.}' & 0.0281 \\
4 & `\emph{a sketch of a \{\}.}' & 0.0265 \\
5 & `\emph{a embroidered \{\}.}' & 0.0235 \\
6 & `\emph{a painting of a \{\}.}' & 0.0231 \\
7 & `\emph{itap of my \{\}.}' & 0.0223 \\
8 & `\emph{a doodle of a \{\}.}' & 0.0207 \\
9 & `\emph{a painting of the \{\}.}' & 0.0202 \\
10 & `\emph{itap of the \{\}.}' & 0.0195 \\
\multicolumn{3}{c}{\vdots} \\
238 & `\emph{there are \{\} shapes in the image.}' & 0.0002 \\
239 & `\emph{the nearest shape in this image is \{\}.}' & 0.0002 \\
240 & `\emph{a zoomed in photo of a "\{\}" traffic sign.}' & 0.0002 \\
241 & `\emph{they look \{\}.}' & 0.0002 \\
242 & `\emph{there are \{\} objects in the image.}' & 0.0002 \\
243 & `\emph{a photo i took while visiting \{\}.}' & 0.0002 \\
244 & `\emph{a fundus image with signs of \{\}}' & 0.0002 \\
245 & `\emph{something rotated at \{\}}' & 0.0001 \\
246 & `\emph{a video of the person during \{\}.}' & 0.0001 \\
247 & `\emph{a photo of the person during \{\}.}' & 0.0001 \\
\cmidrule{1-3}
\multicolumn{3}{c}{\textsc{ImageNet-Sketch}}  \\ 
\cmidrule{1-3}
1 & `\emph{a drawing of a \{\}.}' & 0.1039 \\
2 & `\emph{a drawing of the \{\}.}' & 0.0783 \\
3 & `\emph{a sketch of a \{\}.}' & 0.0763 \\
4 & `\emph{a sketch of the \{\}.}' & 0.0564 \\
5 & `\emph{a black and white photo of a \{\}.}' & 0.0525 \\
6 & `\emph{a doodle of a \{\}.}' & 0.0352 \\
7 & `\emph{a black and white photo of the \{\}.}' & 0.0334 \\
8 & `\emph{a rendering of a \{\}.}' & 0.0249 \\
9 & `\emph{a doodle of the \{\}.}' & 0.0245 \\
10 & `\emph{a rendering of the \{\}.}' & 0.0151 \\
\multicolumn{3}{c}{\vdots} \\
238 & `\emph{something rotated at \{\}}' & 0.0002 \\
239 & `\emph{a photo from my visit to \{\}.}' & 0.0002 \\
240 & `\emph{a photo i took in \{\}.}' & 0.0001 \\
241 & `\emph{they look \{\}.}' & 0.0001 \\
242 & `\emph{there are \{\} objects in the image.}' & 0.0001 \\
243 & `\emph{a fundus image with signs of \{\}}' & 0.0001 \\
244 & `\emph{a video of the person during \{\}.}' & 0.0001 \\
245 & `\emph{a photo of the person during \{\}.}' & 0.0001 \\
246 & `\emph{a zoomed in photo of a "\{\}" traffic sign.}' & 0.0001 \\
247 & `\emph{a photo i took while visiting \{\}.}' & 0.0001 \\
\cmidrule{1-3}
\multicolumn{3}{c}{\textsc{ImageNet-V2}}  \\ 
\cmidrule{1-3}
1 & `\emph{itap of a \{\}.}' & 0.0832 \\
2 & `\emph{itap of the \{\}.}' & 0.0562 \\
3 & `\emph{itap of my \{\}.}' & 0.0533 \\
4 & `\emph{a black and white photo of a \{\}.}' & 0.0207 \\
5 & `\emph{a high contrast photo of a \{\}.}' & 0.0185 \\
6 & `\emph{a photo of a large \{\}.}' & 0.0155 \\
7 & `\emph{a low contrast photo of a \{\}.}' & 0.0152 \\
8 & `\emph{a black and white photo of the \{\}.}' & 0.0148 \\
9 & `\emph{a photo of the large \{\}.}' & 0.0148 \\
10 & `\emph{a bright photo of a \{\}.}' & 0.0133 \\
\multicolumn{3}{c}{\vdots} \\
238 & `\emph{something rotated at \{\}}' & 0.0002 \\
239 & `\emph{the closest shape in this rendered image is \{\}.}' & 0.0002 \\
240 & `\emph{they look \{\}.}' & 0.0001 \\
241 & `\emph{the nearest shape in this image is \{\}.}' & 0.0001 \\
242 & `\emph{there are \{\} shapes in the image.}' & 0.0001 \\
243 & `\emph{a video of the person during \{\}.}' & 0.0001 \\
244 & `\emph{a fundus image with signs of \{\}}' & 0.0001 \\
245 & `\emph{a zoomed in photo of a "\{\}" traffic sign.}' & 0.0001 \\
246 & `\emph{a photo of the person during \{\}.}' & 0.0001 \\
247 & `\emph{there are \{\} objects in the image.}' & 0.0001 \\
\cmidrule{1-3}
\multicolumn{3}{c}{\textsc{Caltech101}}  \\ 
\cmidrule{1-3}
1 & `\emph{a black and white photo of a \{\}.}' & 0.0297 \\
2 & `\emph{itap of a \{\}.}' & 0.0216 \\
3 & `\emph{itap of my \{\}.}' & 0.0201 \\
4 & `\emph{a photo of a large \{\}.}' & 0.0195 \\
5 & `\emph{a black and white photo of the \{\}.}' & 0.0175 \\
6 & `\emph{a photo of the large \{\}.}' & 0.0153 \\
7 & `\emph{a high contrast photo of a \{\}.}' & 0.0150 \\
8 & `\emph{a dark photo of a \{\}.}' & 0.0138 \\
9 & `\emph{a low contrast photo of a \{\}.}' & 0.0128 \\
10 & `\emph{itap of the \{\}.}' & 0.0118 \\
\multicolumn{3}{c}{\vdots} \\
238 & `\emph{a photo i took in \{\}.}' & 0.0003 \\
239 & `\emph{a face that looks \{\}.}' & 0.0003 \\
240 & `\emph{a photo i took while visiting \{\}.}' & 0.0003 \\
241 & `\emph{a fundus image with signs of \{\}}' & 0.0002 \\
242 & `\emph{something rotated at \{\}}' & 0.0002 \\
243 & `\emph{there are \{\} objects in the image.}' & 0.0002 \\
244 & `\emph{a zoomed in photo of a "\{\}" traffic sign.}' & 0.0002 \\
245 & `\emph{a video of the person during \{\}.}' & 0.0002 \\
246 & `\emph{a photo of the person during \{\}.}' & 0.0002 \\
247 & `\emph{they look \{\}.}' & 0.0001 \\
\cmidrule{1-3}
\multicolumn{3}{c}{\textsc{Cars196}}  \\ 
\cmidrule{1-3}
1 & `\emph{a bright photo of the \{\}.}' & 0.0307 \\
2 & `\emph{a bright photo of a \{\}.}' & 0.0223 \\
3 & `\emph{a high contrast photo of the \{\}.}' & 0.0210 \\
4 & `\emph{a photo of the large \{\}.}' & 0.0204 \\
5 & `\emph{a example of \{\}.}' & 0.0203 \\
6 & `\emph{an example of \{\}}' & 0.0179 \\
7 & `\emph{a photo of the clean \{\}.}' & 0.0170 \\
8 & `\emph{a photo of the big \{\}.}' & 0.0167 \\
9 & `\emph{a photo of the \{\}.}' & 0.0166 \\
10 & `\emph{a \{\}.}' & 0.0166 \\
\multicolumn{3}{c}{\vdots} \\
238 & `\emph{a retinal image with \{\}}' & 0.0002 \\
239 & `\emph{an object rotated at \{\}}' & 0.0002 \\
240 & `\emph{a photo of the person during \{\}.}' & 0.0002 \\
241 & `\emph{patient's pathology examination indicates \{\}}' & 0.0002 \\
242 & `\emph{something at a \{\} rotation}' & 0.0001 \\
243 & `\emph{the closest shape in this rendered image is \{\}.}' & 0.0001 \\
244 & `\emph{a zoomed in photo of a "\{\}" traffic sign.}' & 0.0001 \\
245 & `\emph{the nearest shape in this image is \{\}.}' & 0.0001 \\
246 & `\emph{a fundus image with signs of \{\}}' & 0.0001 \\
247 & `\emph{the closest shape in this image is \{\}.}' & 0.0000 \\
\cmidrule{1-3}
\multicolumn{3}{c}{\textsc{CIFAR10}}  \\ 
\cmidrule{1-3}
1 & `\emph{a pixelated photo of a \{\}.}' & 0.0237 \\
2 & `\emph{itap of a \{\}.}' & 0.0152 \\
3 & `\emph{satellite view of a \{\}.}' & 0.0152 \\
4 & `\emph{a rendered image of \{\} shapes.}' & 0.0148 \\
5 & `\emph{a jpeg corrupted photo of a \{\}.}' & 0.0144 \\
6 & `\emph{a \{\} in a video game.}' & 0.0132 \\
7 & `\emph{a pixelated photo of the \{\}.}' & 0.0130 \\
8 & `\emph{a centered satellite photo of a \{\}.}' & 0.0128 \\
9 & `\emph{aerial photo of a \{\}.}' & 0.0122 \\
10 & `\emph{satellite photo of a \{\}.}' & 0.0121 \\
\multicolumn{3}{c}{\vdots} \\
238 & `\emph{a photo i took in \{\}.}' & 0.0008 \\
239 & `\emph{a example of a person during \{\}.}' & 0.0008 \\
240 & `\emph{something at a \{\} rotation}' & 0.0008 \\
241 & `\emph{a demonstration of the person during \{\}.}' & 0.0007 \\
242 & `\emph{a face that looks \{\}.}' & 0.0007 \\
243 & `\emph{a photo from my visit to \{\}.}' & 0.0005 \\
244 & `\emph{a example of the person during \{\}.}' & 0.0005 \\
245 & `\emph{a photo i took while visiting \{\}.}' & 0.0005 \\
246 & `\emph{a photo of the person during \{\}.}' & 0.0003 \\
247 & `\emph{they look \{\}.}' & 0.0003 \\
\cmidrule{1-3}
\multicolumn{3}{c}{\textsc{CIFAR100}}  \\ 
\cmidrule{1-3}
1 & `\emph{a pixelated photo of a \{\}.}' & 0.0270 \\
2 & `\emph{itap of a \{\}.}' & 0.0218 \\
3 & `\emph{satellite view of a \{\}.}' & 0.0210 \\
4 & `\emph{a jpeg corrupted photo of a \{\}.}' & 0.0161 \\
5 & `\emph{a centered satellite photo of a \{\}.}' & 0.0154 \\
6 & `\emph{aerial photo of a \{\}.}' & 0.0143 \\
7 & `\emph{a drawing of a \{\}.}' & 0.0139 \\
8 & `\emph{satellite photo of a \{\}.}' & 0.0130 \\
9 & `\emph{a rendered image of \{\} shapes.}' & 0.0127 \\
10 & `\emph{a pixelated photo of the \{\}.}' & 0.0124 \\
\multicolumn{3}{c}{\vdots} \\
238 & `\emph{a example of a person during \{\}.}' & 0.0005 \\
239 & `\emph{a photo of a person during \{\}.}' & 0.0005 \\
240 & `\emph{a photo i took in \{\}.}' & 0.0004 \\
241 & `\emph{a photo from my visit to \{\}.}' & 0.0004 \\
242 & `\emph{a demonstration of the person during \{\}.}' & 0.0004 \\
243 & `\emph{a video of the person during \{\}.}' & 0.0004 \\
244 & `\emph{they look \{\}.}' & 0.0004 \\
245 & `\emph{a photo i took while visiting \{\}.}' & 0.0003 \\
246 & `\emph{a example of the person during \{\}.}' & 0.0003 \\
247 & `\emph{a photo of the person during \{\}.}' & 0.0001 \\
\cmidrule{1-3}
\multicolumn{3}{c}{\textsc{DTD}}  \\ 
\cmidrule{1-3}
1 & `\emph{itap of a \{\}.}' & 0.0625 \\
2 & `\emph{itap of my \{\}.}' & 0.0336 \\
3 & `\emph{itap of the \{\}.}' & 0.0309 \\
4 & `\emph{a close-up photo of a \{\}.}' & 0.0163 \\
5 & `\emph{aerial view of a \{\}.}' & 0.0121 \\
6 & `\emph{satellite view of a \{\}.}' & 0.0119 \\
7 & `\emph{a photo of a large \{\}.}' & 0.0118 \\
8 & `\emph{a black and white photo of a \{\}.}' & 0.0115 \\
9 & `\emph{a high contrast photo of a \{\}.}' & 0.0107 \\
10 & `\emph{aerial photo of a \{\}.}' & 0.0104 \\
\multicolumn{3}{c}{\vdots} \\
238 & `\emph{something rotated at \{\}}' & 0.0006 \\
239 & `\emph{there are \{\} shapes in the image.}' & 0.0005 \\
240 & `\emph{a video of the person during \{\}.}' & 0.0004 \\
241 & `\emph{an outdoor number \{\} written on a sign}' & 0.0004 \\
242 & `\emph{there are \{\} objects in the image.}' & 0.0004 \\
243 & `\emph{patient's pathology examination indicates \{\}}' & 0.0004 \\
244 & `\emph{a photo of the person during \{\}.}' & 0.0004 \\
245 & `\emph{the number \{\} in the center of the image}' & 0.0003 \\
246 & `\emph{a \{\} review of a movie.}' & 0.0002 \\
247 & `\emph{a fundus image with signs of \{\}}' & 0.0002 \\
\cmidrule{1-3}
\multicolumn{3}{c}{\textsc{EuroSat}}  \\ 
\cmidrule{1-3}
1 & `\emph{satellite view of the \{\}.}' & 0.0548 \\
2 & `\emph{a centered satellite photo of a \{\}.}' & 0.0545 \\
3 & `\emph{satellite view of a \{\}.}' & 0.0529 \\
4 & `\emph{a centered satellite photo of the \{\}.}' & 0.0520 \\
5 & `\emph{a satellite image of \{\}.}' & 0.0495 \\
6 & `\emph{satellite view of \{\}.}' & 0.0456 \\
7 & `\emph{a centered satellite photo of \{\}.}' & 0.0455 \\
8 & `\emph{satellite photo of the \{\}.}' & 0.0418 \\
9 & `\emph{satellite photo of a \{\}.}' & 0.0355 \\
10 & `\emph{a satellite photo of \{\}.}' & 0.0340 \\
\multicolumn{3}{c}{\vdots} \\
238 & `\emph{a demonstration of a person during \{\}.}' & 0.0002 \\
239 & `\emph{a example of a person during \{\}.}' & 0.0002 \\
240 & `\emph{a video of the person during \{\}.}' & 0.0002 \\
241 & `\emph{a example of the person during \{\}.}' & 0.0002 \\
242 & `\emph{a face that looks \{\}.}' & 0.0002 \\
243 & `\emph{a photo of the person during \{\}.}' & 0.0002 \\
244 & `\emph{a photo i took in \{\}.}' & 0.0002 \\
245 & `\emph{a photo i took while visiting \{\}.}' & 0.0002 \\
246 & `\emph{a demonstration of the person during \{\}.}' & 0.0001 \\
247 & `\emph{patient's pathology examination indicates \{\}}' & 0.0001 \\
\cmidrule{1-3}
\multicolumn{3}{c}{\textsc{Food101}}  \\ 
\cmidrule{1-3}
1 & `\emph{itap of a \{\}.}' & 0.0514 \\
2 & `\emph{itap of my \{\}.}' & 0.0490 \\
3 & `\emph{itap of the \{\}.}' & 0.0408 \\
4 & `\emph{a photo of \{\}, a type of food.}' & 0.0244 \\
5 & `\emph{a photo of the large \{\}.}' & 0.0204 \\
6 & `\emph{a example of \{\}.}' & 0.0169 \\
7 & `\emph{a photo of a large \{\}.}' & 0.0154 \\
8 & `\emph{a black and white photo of a \{\}.}' & 0.0145 \\
9 & `\emph{a photo of a big \{\}.}' & 0.0144 \\
10 & `\emph{a low contrast photo of a \{\}.}' & 0.0139 \\
\multicolumn{3}{c}{\vdots} \\
238 & `\emph{a street sign with the number \{\}}' & 0.0001 \\
239 & `\emph{a face that looks \{\}.}' & 0.0001 \\
240 & `\emph{there are \{\} shapes in the image.}' & 0.0001 \\
241 & `\emph{something at a \{\} rotation}' & 0.0001 \\
242 & `\emph{a video of the person during \{\}.}' & 0.0001 \\
243 & `\emph{there are \{\} objects in the image.}' & 0.0001 \\
244 & `\emph{something rotated at \{\}}' & 0.0001 \\
245 & `\emph{a zoomed in photo of a "\{\}" traffic sign.}' & 0.0001 \\
246 & `\emph{a photo of the person during \{\}.}' & 0.0001 \\
247 & `\emph{a fundus image with signs of \{\}}' & 0.0000 \\
\cmidrule{1-3}
\multicolumn{3}{c}{\textsc{Oxford Flowers}}  \\
\cmidrule{1-3}
1 & `\emph{itap of my \{\}.}' & 0.0544 \\
2 & `\emph{itap of a \{\}.}' & 0.0520 \\
3 & `\emph{itap of the \{\}.}' & 0.0415 \\
4 & `\emph{a bright photo of a \{\}.}' & 0.0260 \\
5 & `\emph{a photo of a \{\}, a type of flower.}' & 0.0241 \\
6 & `\emph{a close-up photo of a \{\}.}' & 0.0226 \\
7 & `\emph{a cropped photo of a \{\}.}' & 0.0197 \\
8 & `\emph{a bright photo of the \{\}.}' & 0.0174 \\
9 & `\emph{a low contrast photo of a \{\}.}' & 0.0170 \\
10 & `\emph{a good photo of a \{\}.}' & 0.0163 \\
\multicolumn{3}{c}{\vdots} \\
238 & `\emph{a demonstration of the person during \{\}.}' & 0.0001 \\
239 & `\emph{a video of a person during \{\}.}' & 0.0001 \\
240 & `\emph{a photo of a person during \{\}.}' & 0.0001 \\
241 & `\emph{there are \{\} objects in the image.}' & 0.0001 \\
242 & `\emph{a video of the person performing \{\}.}' & 0.0001 \\
243 & `\emph{a photo of the person performing \{\}.}' & 0.0001 \\
244 & `\emph{a example of the person during \{\}.}' & 0.0000 \\
245 & `\emph{a \{\} review of a movie.}' & 0.0000 \\
246 & `\emph{a video of the person during \{\}.}' & 0.0000 \\
247 & `\emph{a photo of the person during \{\}.}' & 0.0000 \\
\cmidrule{1-3}
\multicolumn{3}{c}{\textsc{Oxford Pets}}  \\ 
\cmidrule{1-3}
1 & `\emph{itap of my \{\}.}' & 0.0450 \\
2 & `\emph{itap of a \{\}.}' & 0.0398 \\
3 & `\emph{a photo of a \{\}, a type of pet.}' & 0.0256 \\
4 & `\emph{itap of the \{\}.}' & 0.0245 \\
5 & `\emph{a bright photo of a \{\}.}' & 0.0233 \\
6 & `\emph{a high contrast photo of a \{\}.}' & 0.0226 \\
7 & `\emph{a photo of a clean \{\}.}' & 0.0202 \\
8 & `\emph{a black and white photo of a \{\}.}' & 0.0199 \\
9 & `\emph{a low contrast photo of a \{\}.}' & 0.0187 \\
10 & `\emph{a example of a person \{\}.}' & 0.0177 \\
\multicolumn{3}{c}{\vdots} \\
238 & `\emph{the nearest shape in this image is \{\}.}' & 0.0001 \\
239 & `\emph{a rendered image of \{\} objects.}' & 0.0000 \\
240 & `\emph{they look \{\}.}' & 0.0000 \\
241 & `\emph{a fundus image with signs of \{\}}' & 0.0000 \\
242 & `\emph{the closest shape in this image is \{\}.}' & 0.0000 \\
243 & `\emph{there are \{\} shapes in the image.}' & 0.0000 \\
244 & `\emph{a photo of the person during \{\}.}' & 0.0000 \\
245 & `\emph{the closest shape in this rendered image is \{\}.}' & 0.0000 \\
246 & `\emph{a video of the person during \{\}.}' & 0.0000 \\
247 & `\emph{there are \{\} objects in the image.}' & 0.0000 \\
\cmidrule{1-3}
\multicolumn{3}{c}{\textsc{Resisc45}}  \\ 
\cmidrule{1-3}
1 & `\emph{satellite view of a \{\}.}' & 0.0767 \\
2 & `\emph{satellite view of the \{\}.}' & 0.0735 \\
3 & `\emph{satellite view of \{\}.}' & 0.0704 \\
4 & `\emph{satellite photo of a \{\}.}' & 0.0524 \\
5 & `\emph{satellite photo of the \{\}.}' & 0.0501 \\
6 & `\emph{a centered satellite photo of a \{\}.}' & 0.0482 \\
7 & `\emph{a satellite image of \{\}.}' & 0.0420 \\
8 & `\emph{satellite photo of \{\}.}' & 0.0420 \\
9 & `\emph{a centered satellite photo of the \{\}.}' & 0.0415 \\
10 & `\emph{a satellite photo of \{\}.}' & 0.0407 \\
\multicolumn{3}{c}{\vdots} \\
238 & `\emph{a close up photo of a "\{\}" traffic sign.}' & 0.0001 \\
239 & `\emph{a photo of the person performing \{\}.}' & 0.0001 \\
240 & `\emph{a demonstration of the person during \{\}.}' & 0.0001 \\
241 & `\emph{a example of a person during \{\}.}' & 0.0001 \\
242 & `\emph{a video of the person during \{\}.}' & 0.0001 \\
243 & `\emph{a fundus image with signs of \{\}}' & 0.0001 \\
244 & `\emph{a example of the person during \{\}.}' & 0.0001 \\
245 & `\emph{a photo of the person during \{\}.}' & 0.0001 \\
246 & `\emph{patient's pathology examination indicates \{\}}' & 0.0001 \\
247 & `\emph{a \{\} review of a movie.}' & 0.0001 \\
\cmidrule{1-3}
\multicolumn{3}{c}{\textsc{Sun397}}  \\ 
\cmidrule{1-3}
1 & `\emph{itap of a \{\}.}' & 0.0320 \\
2 & `\emph{a photo of the large \{\}.}' & 0.0307 \\
3 & `\emph{a photo of a large \{\}.}' & 0.0286 \\
4 & `\emph{a black and white photo of a \{\}.}' & 0.0266 \\
5 & `\emph{itap of the \{\}.}' & 0.0220 \\
6 & `\emph{itap of my \{\}.}' & 0.0195 \\
7 & `\emph{a black and white photo of the \{\}.}' & 0.0176 \\
8 & `\emph{a photo of the small \{\}.}' & 0.0176 \\
9 & `\emph{a bright photo of a \{\}.}' & 0.0168 \\
10 & `\emph{a high contrast photo of a \{\}.}' & 0.0168 \\
\multicolumn{3}{c}{\vdots} \\
238 & `\emph{a face that looks \{\}.}' & 0.0001 \\
239 & `\emph{the closest shape in this rendered image is \{\}.}' & 0.0001 \\
240 & `\emph{patient's pathology examination indicates \{\}}' & 0.0001 \\
241 & `\emph{a video of the person during \{\}.}' & 0.0001 \\
242 & `\emph{something rotated at \{\}}' & 0.0001 \\
243 & `\emph{the closest shape in this image is \{\}.}' & 0.0001 \\
244 & `\emph{the nearest shape in this image is \{\}.}' & 0.0001 \\
245 & `\emph{a close up photo of a "\{\}" traffic sign.}' & 0.0001 \\
246 & `\emph{a fundus image with signs of \{\}}' & 0.0000 \\
247 & `\emph{a zoomed in photo of a "\{\}" traffic sign.}' & 0.0000 \\
\bottomrule
\end{supertabular}
\normalsize

\onecolumn

\section{The Prompt Pool} \label{sec:prompt_pool}

For our experiments in the main text and all of our ablations except \cref{sec:gpt_prompts}, we constructed our pool of prompts taking the union of all of the prompts from \citet{radford2021learning}---which can be found at \url{https://github.com/openai/CLIP/blob/main/data/prompts.md} and \url{https://github.com/openai/CLIP/blob/main/notebooks/Prompt_Engineering_for_ImageNet.ipynb}, for non-ImageNet and ImageNet, respectively---and \citet{zhai2021lit}---which can be found Tables 11 and 12 of their paper.

The resulting pool of 247 prompts is as follows:

\mdfsetup{
backgroundcolor=mplblue!3,
middlelinecolor=mplblue!75!white,
roundcorner=8pt,
middlelinewidth=2
}
\begin{mdframed}
\fontsize{7.5}{8}\selectfont
\emph{satellite imagery of \{\}.}$\,\cdot\,$
\emph{aerial view of a \{\}.}$\,\cdot\,$
\emph{i love my \{\}!}$\,\cdot\,$
\emph{a drawing of the \{\}.}$\,\cdot\,$
\emph{a video of a person \{\}.}$\,\cdot\,$
\emph{satellite photo of \{\}.}$\,\cdot\,$
\emph{a photo of the person performing \{\}.}$\,\cdot\,$
\emph{there are \{\} shapes.}$\,\cdot\,$
\emph{a video of the person using \{\}.}$\,\cdot\,$
\emph{a centered satellite photo of \{\}.}$\,\cdot\,$
\emph{a example of the person doing \{\}.}$\,\cdot\,$
\emph{a photo of a person practicing \{\}.}$\,\cdot\,$
\emph{a example of the person performing \{\}.}$\,\cdot\,$
\emph{art of a \{\}.}$\,\cdot\,$
\emph{a \{\}.}$\,\cdot\,$
\emph{itap of the \{\}.}$\,\cdot\,$
\emph{a drawing of a \{\}.}$\,\cdot\,$
\emph{a origami \{\}.}$\,\cdot\,$
\emph{a video of \{\}.}$\,\cdot\,$
\emph{a photo of a nice \{\}.}$\,\cdot\,$
\emph{a blurry photo of a \{\}.}$\,\cdot\,$
\emph{they look \{\}.}$\,\cdot\,$
\emph{the \{\} in a video game.}$\,\cdot\,$
\emph{a face that looks \{\}.}$\,\cdot\,$
\emph{a picture of \{\} objects.}$\,\cdot\,$
\emph{a close-up photo of the \{\}.}$\,\cdot\,$
\emph{a photo of \{\}.}$\,\cdot\,$
\emph{a photo i took in \{\}.}$\,\cdot\,$
\emph{a example of the person during \{\}.}$\,\cdot\,$
\emph{a centered satellite photo of the \{\}.}$\,\cdot\,$
\emph{a street sign with the number \{\}}$\,\cdot\,$
\emph{a photo of a clean \{\}.}$\,\cdot\,$
\emph{a photo of a weird \{\}.}$\,\cdot\,$
\emph{a photo of a small \{\}.}$\,\cdot\,$
\emph{a high contrast photo of a \{\}.}$\,\cdot\,$
\emph{the nearest shape in this image is \{\}.}$\,\cdot\,$
\emph{a photo of the large \{\}.}$\,\cdot\,$
\emph{an example of \{\}}$\,\cdot\,$
\emph{a pixelated photo of the \{\}.}$\,\cdot\,$
\emph{a histopathology slide showing \{\}}$\,\cdot\,$
\emph{a embroidered \{\}.}$\,\cdot\,$
\emph{satellite view of a \{\}.}$\,\cdot\,$
\emph{a high contrast photo of the \{\}.}$\,\cdot\,$
\emph{a photo of the \{\} texture.}$\,\cdot\,$
\emph{the closest shape in this rendered image is \{\}.}$\,\cdot\,$
\emph{a \{\} slide}$\,\cdot\,$
\emph{a demonstration of a person doing \{\}.}$\,\cdot\,$
\emph{a demonstration of a person practicing \{\}.}$\,\cdot\,$
\emph{this is a photo of \{\}}$\,\cdot\,$
\emph{a demonstration of the person using \{\}.}$\,\cdot\,$
\emph{a example of the person using \{\}.}$\,\cdot\,$
\emph{a photo of the person doing \{\}.}$\,\cdot\,$
\emph{a video of the person during \{\}.}$\,\cdot\,$
\emph{the number \{\} in the center of the image}$\,\cdot\,$
\emph{an example histopathological image showing \{\}}$\,\cdot\,$
\emph{a photo of the clean \{\}.}$\,\cdot\,$
\emph{a demonstration of the person practicing \{\}.}$\,\cdot\,$
\emph{the origami \{\}.}$\,\cdot\,$
\emph{the plushie \{\}.}$\,\cdot\,$
\emph{a photo of a \{\} thing.}$\,\cdot\,$
\emph{a photo of a cool \{\}.}$\,\cdot\,$
\emph{a sculpture of the \{\}.}$\,\cdot\,$
\emph{a example of a person during \{\}.}$\,\cdot\,$
\emph{a demonstration of the person \{\}.}$\,\cdot\,$
\emph{a low resolution photo of the \{\}.}$\,\cdot\,$
\emph{look at how \{\} they are.}$\,\cdot\,$
\emph{a photo of a person doing \{\}.}$\,\cdot\,$
\emph{a photo of the \{\} pattern.}$\,\cdot\,$
\emph{a bad photo of the \{\}.}$\,\cdot\,$
\emph{a \{\} texture}$\,\cdot\,$
\emph{the number \{\}}$\,\cdot\,$
\emph{aerial imagery of \{\}.}$\,\cdot\,$
\emph{a photo of a person \{\}.}$\,\cdot\,$
\emph{a jpeg corrupted photo of a \{\}.}$\,\cdot\,$
\emph{\{\} objects}$\,\cdot\,$
\emph{a photo of \{\} objects.}$\,\cdot\,$
\emph{a \{\} flower}$\,\cdot\,$
\emph{a rendition of the \{\}.}$\,\cdot\,$
\emph{a photo of the cool \{\}.}$\,\cdot\,$
\emph{\{\}}$\,\cdot\,$
\emph{a low resolution photo of a \{\}.}$\,\cdot\,$
\emph{\{\} shapes}$\,\cdot\,$
\emph{a photo from my home country of \{\}.}$\,\cdot\,$
\emph{a cropped photo of the \{\}.}$\,\cdot\,$
\emph{the plastic \{\}.}$\,\cdot\,$
\emph{a sculpture of a \{\}.}$\,\cdot\,$
\emph{a pixelated photo of a \{\}.}$\,\cdot\,$
\emph{itap of a \{\}.}$\,\cdot\,$
\emph{a demonstration of \{\}.}$\,\cdot\,$
\emph{a video of a person using \{\}.}$\,\cdot\,$
\emph{a doodle of a \{\}.}$\,\cdot\,$
\emph{a photo of the \{\} object.}$\,\cdot\,$
\emph{a sketch of a \{\}.}$\,\cdot\,$
\emph{a \{\} plant}$\,\cdot\,$
\emph{a satellite image of \{\}.}$\,\cdot\,$
\emph{a plastic \{\}.}$\,\cdot\,$
\emph{\{\} thing}$\,\cdot\,$
\emph{\{\} things}$\,\cdot\,$
\emph{a photo of the person using \{\}.}$\,\cdot\,$
\emph{itap of my \{\}.}$\,\cdot\,$
\emph{a example of a person using \{\}.}$\,\cdot\,$
\emph{the closest shape in this image is \{\}.}$\,\cdot\,$
\emph{a close-up photo of a \{\}.}$\,\cdot\,$
\emph{a bright photo of a \{\}.}$\,\cdot\,$
\emph{a photo of the person during \{\}.}$\,\cdot\,$
\emph{art of the \{\}.}$\,\cdot\,$
\emph{graffiti of the \{\}.}$\,\cdot\,$
\emph{a tattoo of a \{\}.}$\,\cdot\,$
\emph{a video of the person performing \{\}.}$\,\cdot\,$
\emph{a photo of a face looking \{\}.}$\,\cdot\,$
\emph{a sketch of the \{\}.}$\,\cdot\,$
\emph{aerial imagery of the \{\}.}$\,\cdot\,$
\emph{a dark photo of a \{\}.}$\,\cdot\,$
\emph{a tattoo of the \{\}.}$\,\cdot\,$
\emph{there are \{\} objects in the image.}$\,\cdot\,$
\emph{\{\}, an animal}$\,\cdot\,$
\emph{a photo of the dirty \{\}.}$\,\cdot\,$
\emph{a example of a person performing \{\}.}$\,\cdot\,$
\emph{a centered photo of a "\{\}" traffic sign.}$\,\cdot\,$
\emph{a photo of the number: "\{\}".}$\,\cdot\,$
\emph{an overhead view of \{\}.}$\,\cdot\,$
\emph{a black and white photo of the \{\}.}$\,\cdot\,$
\emph{a zoomed in photo of a "\{\}" traffic sign.}$\,\cdot\,$
\emph{a example of \{\}.}$\,\cdot\,$
\emph{a photo of a \{\}.}$\,\cdot\,$
\emph{a retinal image with \{\}}$\,\cdot\,$
\emph{a photo of the \{\}, a type of aircraft.}$\,\cdot\,$
\emph{a photo of a \{\} texture.}$\,\cdot\,$
\emph{a demonstration of a person during \{\}.}$\,\cdot\,$
\emph{a \{\} texture.}$\,\cdot\,$
\emph{a \{\} in a video game.}$\,\cdot\,$
\emph{a painting of the \{\}.}$\,\cdot\,$
\emph{a cropped photo of a \{\}.}$\,\cdot\,$
\emph{a demonstration of the person doing \{\}.}$\,\cdot\,$
\emph{a photo of a \{\} pattern.}$\,\cdot\,$
\emph{a example of a person practicing \{\}.}$\,\cdot\,$
\emph{a photo of a large \{\}.}$\,\cdot\,$
\emph{a photo from my visit to \{\}.}$\,\cdot\,$
\emph{an overhead image of \{\}.}$\,\cdot\,$
\emph{a photo of the weird \{\}.}$\,\cdot\,$
\emph{aerial photo of \{\}.}$\,\cdot\,$
\emph{satellite imagery of the \{\}.}$\,\cdot\,$
\emph{graffiti of a \{\}.}$\,\cdot\,$
\emph{a close up photo of a "\{\}" traffic sign.}$\,\cdot\,$
\emph{a photo of a \{\}, a type of pet.}$\,\cdot\,$
\emph{a low contrast photo of a \{\}.}$\,\cdot\,$
\emph{a satellite photo of \{\}.}$\,\cdot\,$
\emph{a video of a person practicing \{\}.}$\,\cdot\,$
\emph{a demonstration of a person using \{\}.}$\,\cdot\,$
\emph{a painting of a \{\}.}$\,\cdot\,$
\emph{a cartoon \{\}.}$\,\cdot\,$
\emph{a photo of my new \{\}.}$\,\cdot\,$
\emph{aerial imagery of a \{\}.}$\,\cdot\,$
\emph{the cartoon \{\}.}$\,\cdot\,$
\emph{a low contrast photo of the \{\}.}$\,\cdot\,$
\emph{a photo of the big \{\}.}$\,\cdot\,$
\emph{a type of pet \{\}}$\,\cdot\,$
\emph{a video of the person \{\}.}$\,\cdot\,$
\emph{a video of a person performing \{\}.}$\,\cdot\,$
\emph{aerial view of the \{\}.}$\,\cdot\,$
\emph{a photo of a person during \{\}.}$\,\cdot\,$
\emph{a photo of a \{\}, a type of aircraft.}$\,\cdot\,$
\emph{a video of a person during \{\}.}$\,\cdot\,$
\emph{a good photo of the \{\}.}$\,\cdot\,$
\emph{a photo of a \{\}, a type of bird.}$\,\cdot\,$
\emph{there are \{\} objects.}$\,\cdot\,$
\emph{a jpeg corrupted photo of the \{\}.}$\,\cdot\,$
\emph{a photo of the \{\} thing.}$\,\cdot\,$
\emph{a photo of a face showing the emotion: \{\}.}$\,\cdot\,$
\emph{a bad photo of a \{\}.}$\,\cdot\,$
\emph{a photo of the small \{\}.}$\,\cdot\,$
\emph{a picture of \{\} shapes.}$\,\cdot\,$
\emph{a centered satellite photo of a \{\}.}$\,\cdot\,$
\emph{a photo of a person using \{\}.}$\,\cdot\,$
\emph{aerial photo of a \{\}.}$\,\cdot\,$
\emph{a photo of a \{\}, a type of flower.}$\,\cdot\,$
\emph{a \{\} review of a movie.}$\,\cdot\,$
\emph{a rendering of the \{\}.}$\,\cdot\,$
\emph{a photo of a dirty \{\}.}$\,\cdot\,$
\emph{satellite imagery of a \{\}.}$\,\cdot\,$
\emph{a rendition of a \{\}.}$\,\cdot\,$
\emph{\{\} rotation}$\,\cdot\,$
\emph{photo of \{\} from the sky.}$\,\cdot\,$
\emph{a blurry photo of the \{\}.}$\,\cdot\,$
\emph{the toy \{\}.}$\,\cdot\,$
\emph{a video of a person doing \{\}.}$\,\cdot\,$
\emph{something at a \{\} rotation}$\,\cdot\,$
\emph{a photo of my clean \{\}.}$\,\cdot\,$
\emph{a example of a person \{\}.}$\,\cdot\,$
\emph{a demonstration of a person performing \{\}.}$\,\cdot\,$
\emph{the embroidered \{\}.}$\,\cdot\,$
\emph{aerial photo of the \{\}.}$\,\cdot\,$
\emph{a video of the person practicing \{\}.}$\,\cdot\,$
\emph{\{\} from above.}$\,\cdot\,$
\emph{a photo of the person practicing \{\}.}$\,\cdot\,$
\emph{a rendering of a \{\}.}$\,\cdot\,$
\emph{there are \{\} shapes in the image.}$\,\cdot\,$
\emph{a photo of a \{\} looking face.}$\,\cdot\,$
\emph{a rendered image of \{\} objects.}$\,\cdot\,$
\emph{an aerial view of \{\}.}$\,\cdot\,$
\emph{a photo of a big \{\}.}$\,\cdot\,$
\emph{a example of a person doing \{\}.}$\,\cdot\,$
\emph{an outdoor house number \{\}}$\,\cdot\,$
\emph{a photo of a hard to see \{\}.}$\,\cdot\,$
\emph{a dark photo of the \{\}.}$\,\cdot\,$
\emph{a example of the person \{\}.}$\,\cdot\,$
\emph{a demonstration of a person \{\}.}$\,\cdot\,$
\emph{a doodle of the \{\}.}$\,\cdot\,$
\emph{a good photo of a \{\}.}$\,\cdot\,$
\emph{an object rotated at \{\}}$\,\cdot\,$
\emph{a photo of the \{\}.}$\,\cdot\,$
\emph{a photo of many \{\}.}$\,\cdot\,$
\emph{a rendered image of \{\} shapes.}$\,\cdot\,$
\emph{histopathology image of \{\}}$\,\cdot\,$
\emph{a plushie \{\}.}$\,\cdot\,$
\emph{a photo i took while visiting \{\}.}$\,\cdot\,$
\emph{patient's pathology examination indicates \{\}}$\,\cdot\,$
\emph{an outdoor number \{\} written on a sign}$\,\cdot\,$
\emph{a photo of the person \{\}.}$\,\cdot\,$
\emph{a photo showing the country of \{\}.}$\,\cdot\,$
\emph{a photo of a person performing \{\}.}$\,\cdot\,$
\emph{a photo of the nice \{\}.}$\,\cdot\,$
\emph{a demonstration of the person during \{\}.}$\,\cdot\,$
\emph{a bright photo of the \{\}.}$\,\cdot\,$
\emph{satellite view of the \{\}.}$\,\cdot\,$
\emph{a example of the person practicing \{\}.}$\,\cdot\,$
\emph{aerial view of \{\}.}$\,\cdot\,$
\emph{a photo of my old \{\}.}$\,\cdot\,$
\emph{a retina with \{\}}$\,\cdot\,$
\emph{a centered image of the number \{\}}$\,\cdot\,$
\emph{a fundus image with signs of \{\}}$\,\cdot\,$
\emph{an object located \{\}}$\,\cdot\,$
\emph{something rotated at \{\}}$\,\cdot\,$
\emph{satellite photo of a \{\}.}$\,\cdot\,$
\emph{a toy \{\}.}$\,\cdot\,$
\emph{a photo of a \{\} object.}$\,\cdot\,$
\emph{a video of the person doing \{\}.}$\,\cdot\,$
\emph{a photo of \{\}, a type of food.}$\,\cdot\,$
\emph{a photo of the hard to see \{\}.}$\,\cdot\,$
\emph{satellite photo of the \{\}.}$\,\cdot\,$
\emph{a photo of one \{\}.}$\,\cdot\,$
\emph{a photo of my dirty \{\}.}$\,\cdot\,$
\emph{a photo of my \{\}.}$\,\cdot\,$
\emph{a photo of the number \{\} written on a sign}$\,\cdot\,$
\emph{satellite view of \{\}.}$\,\cdot\,$
\emph{a demonstration of the person performing \{\}.}$\,\cdot\,$
\emph{a black and white photo of a \{\}.}
\normalsize
\end{mdframed}

For our experiments in \cref{sec:gpt_prompts}, we created an additional 179 unique prompts resulting in a enlarged pool of 426 total prompts. These additional prompts were created with ChatGPT \citep{chatgpt}, in the following conversation. We used the first output from ChatGPT without performing any re-rolls. Note that although ChatGPT produced some duplicate prompts, our final pool contains only unique prompts.

\begin{displayquote}
\fontfamily{lmodern}\selectfont
I want you to act as a template generator. Templates take the form of sentences that describe photographs, with ``\{\}'' symbols that can be replaced with the content of the photographs. I will provide you with the form of the templates. I will use ``XXX'' to indicate the part of the template that you should change. I will also provide you with a few examples. I want you to generate 50 unique templates of the given form. Do not repeat the examples. The first form is ``A photo of a \{\}, a type of XXX.''. Examples are: ``A photo of a \{\}, a type of bird.'', ``A photo of a \{\}, a type of animal.'', ``A photo of a \{\}, a type of flower.''. 
\end{displayquote}

\begin{mdframed}
\fontsize{7.5}{8}\selectfont
\emph{A photo of a \{\}, a type of insect.}$\,\cdot\,$
\emph{A photo of a \{\}, a type of fish.}$\,\cdot\,$
\emph{A photo of a \{\}, a type of tree.}$\,\cdot\,$
\emph{A photo of a \{\}, a type of fruit.}$\,\cdot\,$
\emph{A photo of a \{\}, a type of car.}$\,\cdot\,$
\emph{A photo of a \{\}, a type of dog.}$\,\cdot\,$
\emph{A photo of a \{\}, a type of mammal.}$\,\cdot\,$
\emph{A photo of a \{\}, a type of reptile.}$\,\cdot\,$
\emph{A photo of a \{\}, a type of food.}$\,\cdot\,$
\emph{A photo of a \{\}, a type of vegetable.}$\,\cdot\,$
\emph{A photo of a \{\}, a type of landscape.}$\,\cdot\,$
\emph{A photo of a \{\}, a type of cityscape.}$\,\cdot\,$
\emph{A photo of a \{\}, a type of seascape.}$\,\cdot\,$
\emph{A photo of a \{\}, a type of architecture.}$\,\cdot\,$
\emph{A photo of a \{\}, a type of monument.}$\,\cdot\,$
\emph{A photo of a \{\}, a type of painting.}$\,\cdot\,$
\emph{A photo of a \{\}, a type of sculpture.}$\,\cdot\,$
\emph{A photo of a \{\}, a type of musical instrument.}$\,\cdot\,$
\emph{A photo of a \{\}, a type of weapon.}$\,\cdot\,$
\emph{A photo of a \{\}, a type of clothing.}$\,\cdot\,$
\emph{A photo of a \{\}, a type of jewelry.}$\,\cdot\,$
\emph{A photo of a \{\}, a type of household item.}$\,\cdot\,$
\emph{A photo of a \{\}, a type of electronic device.}$\,\cdot\,$
\emph{A photo of a \{\}, a type of tool.}$\,\cdot\,$
\emph{A photo of a \{\}, a type of transportation.}$\,\cdot\,$
\emph{A photo of a \{\}, a type of recreational activity.}$\,\cdot\,$
\emph{A photo of a \{\}, a type of game.}$\,\cdot\,$
\emph{A photo of a \{\}, a type of sport.}$\,\cdot\,$
\emph{A photo of a \{\}, a type of musical genre.}$\,\cdot\,$
\emph{A photo of a \{\}, a type of movie genre.}$\,\cdot\,$
\emph{A photo of a \{\}, a type of book genre.}$\,\cdot\,$
\emph{A photo of a \{\}, a type of historical event.}$\,\cdot\,$
\emph{A photo of a \{\}, a type of mythological creature.}$\,\cdot\,$
\emph{A photo of a \{\}, a type of fantasy creature.}$\,\cdot\,$
\emph{A photo of a \{\}, a type of planet.}$\,\cdot\,$
\emph{A photo of a \{\}, a type of constellation.}$\,\cdot\,$
\emph{A photo of a \{\}, a type of comet.}$\,\cdot\,$
\emph{A photo of a \{\}, a type of galaxy.}$\,\cdot\,$
\emph{A photo of a \{\}, a type of meteor.}$\,\cdot\,$
\emph{A photo of a \{\}, a type of asteroid.}$\,\cdot\,$
\emph{A photo of a \{\}, a type of planet.}$\,\cdot\,$
\emph{A photo of a \{\}, a type of star.}$\,\cdot\,$
\emph{A photo of a \{\}, a type of black hole.}$\,\cdot\,$
\emph{A photo of a \{\}, a type of neutron star.}$\,\cdot\,$
\emph{A photo of a \{\}, a type of quasar.}$\,\cdot\,$
\emph{A photo of a \{\}, a type of pulsar.}$\,\cdot\,$
\emph{A photo of a \{\}, a type of supernova.}$\,\cdot\,$
\emph{A photo of a \{\}, a type of brown dwarf.}$\,\cdot\,$
\emph{A photo of a \{\}, a type of white dwarf.}$\,\cdot\,$
\emph{A photo of a \{\}, a type of red giant.}
\normalsize
\end{mdframed}

\begin{displayquote}
\fontfamily{lmodern}\selectfont
Generate another 50 for the same template please.
\end{displayquote}

\begin{mdframed}
\fontsize{7.5}{8}\selectfont
\emph{A photo of a \{\}, a type of butterfly.}$\,\cdot\,$
\emph{A photo of a \{\}, a type of amphibian.}$\,\cdot\,$
\emph{A photo of a \{\}, a type of berry.}$\,\cdot\,$
\emph{A photo of a \{\}, a type of motorcycle.}$\,\cdot\,$
\emph{A photo of a \{\}, a type of cat.}$\,\cdot\,$
\emph{A photo of a \{\}, a type of rodent.}$\,\cdot\,$
\emph{A photo of a \{\}, a type of fish.}$\,\cdot\,$
\emph{A photo of a \{\}, a type of dinosaur.}$\,\cdot\,$
\emph{A photo of a \{\}, a type of pasta.}$\,\cdot\,$
\emph{A photo of a \{\}, a type of grain.}$\,\cdot\,$
\emph{A photo of a \{\}, a type of mountain range.}$\,\cdot\,$
\emph{A photo of a \{\}, a type of waterfall.}$\,\cdot\,$
\emph{A photo of a \{\}, a type of lake.}$\,\cdot\,$
\emph{A photo of a \{\}, a type of bridge.}$\,\cdot\,$
\emph{A photo of a \{\}, a type of lighthouse.}$\,\cdot\,$
\emph{A photo of a \{\}, a type of pottery.}$\,\cdot\,$
\emph{A photo of a \{\}, a type of tapestry.}$\,\cdot\,$
\emph{A photo of a \{\}, a type of drum.}$\,\cdot\,$
\emph{A photo of a \{\}, a type of sword.}$\,\cdot\,$
\emph{A photo of a \{\}, a type of hat.}$\,\cdot\,$
\emph{A photo of a \{\}, a type of watch.}$\,\cdot\,$
\emph{A photo of a \{\}, a type of kitchen appliance.}$\,\cdot\,$
\emph{A photo of a \{\}, a type of camera.}$\,\cdot\,$
\emph{A photo of a \{\}, a type of power tool.}$\,\cdot\,$
\emph{A photo of a \{\}, a type of boat.}$\,\cdot\,$
\emph{A photo of a \{\}, a type of adventure sport.}$\,\cdot\,$
\emph{A photo of a \{\}, a type of board game.}$\,\cdot\,$
\emph{A photo of a \{\}, a type of ball sport.}$\,\cdot\,$
\emph{A photo of a \{\}, a type of folk music.}$\,\cdot\,$
\emph{A photo of a \{\}, a type of action movie.}$\,\cdot\,$
\emph{A photo of a \{\}, a type of mystery novel.}$\,\cdot\,$
\emph{A photo of a \{\}, a type of war.}$\,\cdot\,$
\emph{A photo of a \{\}, a type of mythical king.}$\,\cdot\,$
\emph{A photo of a \{\}, a type of fantasy race.}$\,\cdot\,$
\emph{A photo of a \{\}, a type of planet.}$\,\cdot\,$
\emph{A photo of a \{\}, a type of constellation.}$\,\cdot\,$
\emph{A photo of a \{\}, a type of comet.}$\,\cdot\,$
\emph{A photo of a \{\}, a type of galaxy.}$\,\cdot\,$
\emph{A photo of a \{\}, a type of meteor.}$\,\cdot\,$
\emph{A photo of a \{\}, a type of asteroid.}$\,\cdot\,$
\emph{A photo of a \{\}, a type of planet.}$\,\cdot\,$
\emph{A photo of a \{\}, a type of star.}$\,\cdot\,$
\emph{A photo of a \{\}, a type of black hole.}$\,\cdot\,$
\emph{A photo of a \{\}, a type of neutron star.}$\,\cdot\,$
\emph{A photo of a \{\}, a type of quasar.}$\,\cdot\,$
\emph{A photo of a \{\}, a type of pulsar.}$\,\cdot\,$
\emph{A photo of a \{\}, a type of supernova.}$\,\cdot\,$
\emph{A photo of a \{\}, a type of brown dwarf.}$\,\cdot\,$
\emph{A photo of a \{\}, a type of white dwarf.}$\,\cdot\,$
\emph{A photo of a \{\}, a type of red giant.}
\normalsize
\end{mdframed}

\begin{displayquote}
\fontfamily{lmodern}\selectfont
The next template takes the following form: ``A XXX photo of a \{\}.'', where XXX describes the photo. Examples are ``A black-and-white photo of a \{\}.'', ``A blurry photo of a \{\}.'', and ``A pixelated photo of a \{\}.''.
\end{displayquote}

\begin{mdframed}
\fontsize{7.5}{8}\selectfont
\emph{A panoramic photo of a \{\}.}$\,\cdot\,$
\emph{A close-up photo of a \{\}.}$\,\cdot\,$
\emph{A wide-angle photo of a \{\}.}$\,\cdot\,$
\emph{A high-resolution photo of a \{\}.}$\,\cdot\,$
\emph{A low-light photo of a \{\}.}$\,\cdot\,$
\emph{A time-lapse photo of a \{\}.}$\,\cdot\,$
\emph{A long-exposure photo of a \{\}.}$\,\cdot\,$
\emph{A night photo of a \{\}.}$\,\cdot\,$
\emph{A sunset photo of a \{\}.}$\,\cdot\,$
\emph{A sunrise photo of a \{\}.}$\,\cdot\,$
\emph{A silhouette photo of a \{\}.}$\,\cdot\,$
\emph{A sepia-toned photo of a \{\}.}$\,\cdot\,$
\emph{A colored photo of a \{\}.}$\,\cdot\,$
\emph{A watercolor photo of a \{\}.}$\,\cdot\,$
\emph{A sketch photo of a \{\}.}$\,\cdot\,$
\emph{A hyperlapse photo of a \{\}.}$\,\cdot\,$
\emph{A tilt-shift photo of a \{\}.}$\,\cdot\,$
\emph{A motion-blurred photo of a \{\}.}$\,\cdot\,$
\emph{A double-exposure photo of a \{\}.}$\,\cdot\,$
\emph{A HDR photo of a \{\}.}$\,\cdot\,$
\emph{A 360-degree photo of a \{\}.}$\,\cdot\,$
\emph{A black-and-white negative photo of a \{\}.}$\,\cdot\,$
\emph{A split-tone photo of a \{\}.}$\,\cdot\,$
\emph{A film-grain photo of a \{\}.}$\,\cdot\,$
\emph{A thermal photo of a \{\}.}$\,\cdot\,$
\emph{A infrared photo of a \{\}.}$\,\cdot\,$
\emph{A ultraviolet photo of a \{\}.}$\,\cdot\,$
\emph{A x-ray photo of a \{\}.}$\,\cdot\,$
\emph{A 3D photo of a \{\}.}$\,\cdot\,$
\emph{A stop-motion photo of a \{\}.}$\,\cdot\,$
\emph{A bokeh photo of a \{\}.}$\,\cdot\,$
\emph{A miniature photo of a \{\}.}$\,\cdot\,$
\emph{A light-painted photo of a \{\}.}$\,\cdot\,$
\emph{A composite photo of a \{\}.}$\,\cdot\,$
\emph{A polarized photo of a \{\}.}$\,\cdot\,$
\emph{A photomontage photo of a \{\}.}$\,\cdot\,$
\emph{A digital-art photo of a \{\}.}$\,\cdot\,$
\emph{A abstract photo of a \{\}.}$\,\cdot\,$
\emph{A selective-focus photo of a \{\}.}$\,\cdot\,$
\emph{A black-and-white film photo of a \{\}.}$\,\cdot\,$
\emph{A cross-processed photo of a \{\}.}$\,\cdot\,$
\emph{A cyanotype photo of a \{\}.}$\,\cdot\,$
\emph{A lomography photo of a \{\}.}$\,\cdot\,$
\emph{A pinhole photo of a \{\}.}$\,\cdot\,$
\emph{A cyanotype photo of a \{\}.}$\,\cdot\,$
\emph{A high-dynamic-range photo of a \{\}.}$\,\cdot\,$
\emph{A low-dynamic-range photo of a \{\}.}$\,\cdot\,$
\emph{A multiexposure photo of a \{\}.}$\,\cdot\,$
\emph{A high-speed photo of a \{\}.}$\,\cdot\,$
\emph{A underwater photo of a \{\}.}
\normalsize
\end{mdframed}

\begin{displayquote}
\fontfamily{lmodern}\selectfont
The next template takes the following form: ``A XXX of a \{\}.'', where XXX describes a medium for capturing images. Examples are ``A photo of a \{\}.'', ``A video of a \{\}.'', ``A cartoon of a \{\}.'', ``A drawing of a \{\}.'', and a ``A painting of a \{\}''.
\end{displayquote}

\begin{mdframed}
\fontsize{7.5}{8}\selectfont
\emph{A sculpture of a \{\}.}$\,\cdot\,$
\emph{A print of a \{\}.}$\,\cdot\,$
\emph{A sketch of a \{\}.}$\,\cdot\,$
\emph{A engraving of a \{\}.}$\,\cdot\,$
\emph{A etching of a \{\}.}$\,\cdot\,$
\emph{A lithograph of a \{\}.}$\,\cdot\,$
\emph{A watercolor of a \{\}.}$\,\cdot\,$
\emph{A pastel of a \{\}.}$\,\cdot\,$
\emph{A charcoal of a \{\}.}$\,\cdot\,$
\emph{A oil painting of a \{\}.}$\,\cdot\,$
\emph{A acrylic painting of a \{\}.}$\,\cdot\,$
\emph{A digital painting of a \{\}.}$\,\cdot\,$
\emph{A fresco of a \{\}.}$\,\cdot\,$
\emph{A mosaic of a \{\}.}$\,\cdot\,$
\emph{A collage of a \{\}.}$\,\cdot\,$
\emph{A graffiti of a \{\}.}$\,\cdot\,$
\emph{A stained glass of a \{\}.}$\,\cdot\,$
\emph{A quilt of a \{\}.}$\,\cdot\,$
\emph{A tapestry of a \{\}.}$\,\cdot\,$
\emph{A batik of a \{\}.}$\,\cdot\,$
\emph{A calligraphy of a \{\}.}$\,\cdot\,$
\emph{A wood carving of a \{\}.}$\,\cdot\,$
\emph{A metal sculpture of a \{\}.}$\,\cdot\,$
\emph{A glass sculpture of a \{\}.}$\,\cdot\,$
\emph{A clay sculpture of a \{\}.}$\,\cdot\,$
\emph{A ice sculpture of a \{\}.}$\,\cdot\,$
\emph{A sand sculpture of a \{\}.}$\,\cdot\,$
\emph{A paper mache of a \{\}.}$\,\cdot\,$
\emph{A sculptural installation of a \{\}.}$\,\cdot\,$
\emph{A mural of a \{\}.}$\,\cdot\,$
\emph{A fresco of a \{\}.}$\,\cdot\,$
\emph{A graffiti of a \{\}.}$\,\cdot\,$
\emph{A street art of a \{\}.}$\,\cdot\,$
\emph{A digital art of a \{\}.}$\,\cdot\,$
\emph{A film of a \{\}.}$\,\cdot\,$
\emph{A animation of a \{\}.}$\,\cdot\,$
\emph{A stop motion animation of a \{\}.}$\,\cdot\,$
\emph{A motion graphics of a \{\}.}$\,\cdot\,$
\emph{A 3D animation of a \{\}.}$\,\cdot\,$
\emph{A VR of a \{\}.}$\,\cdot\,$
\emph{A AR of a \{\}.}$\,\cdot\,$
\emph{A hologram of a \{\}.}$\,\cdot\,$
\emph{A laser show of a \{\}.}$\,\cdot\,$
\emph{A light show of a \{\}.}$\,\cdot\,$
\emph{A pyrotechnics of a \{\}.}$\,\cdot\,$
\emph{A performance of a \{\}.}$\,\cdot\,$
\emph{A sound sculpture of a \{\}.}$\,\cdot\,$
\emph{A kinetic sculpture of a \{\}.}$\,\cdot\,$
\emph{A land art of a \{\}.}$\,\cdot\,$
\emph{A environmental art of a \{\}.}
\normalsize
\end{mdframed}

\end{document}